\documentclass[11pt]{article}

\usepackage[final]{acl}

\usepackage{times}
\usepackage{latexsym}
\usepackage{booktabs}
\usepackage{pdfpages}
\usepackage{booktabs}
\usepackage{multirow}
\usepackage{amssymb}
\usepackage[table]{xcolor}
\usepackage{tcolorbox}
\usepackage{enumitem}
\usepackage{calc}
\tcbuselibrary{breakable}

\usepackage{tcolorbox}
\tcbuselibrary{breakable, skins}
\usepackage[T1]{fontenc}

\usepackage[utf8]{inputenc}

\usepackage{microtype}

\usepackage{inconsolata}

\usepackage{graphicx}
\usepackage{tcolorbox}
\usepackage{xcolor}
\usepackage{float}        
\usepackage{caption} 
\usepackage{placeins}
\usepackage{cleveref}
\usepackage{url}
\usepackage{xspace}

\newcommand*{\afrifact}{AfrIFact}
\newcommand*{\afrifacthealth}{\textit{AfrIFact-Health}}

\newcommand*{\afrifactculture}{\textit{AfrIFact-Culture-News}}

\newcommand*{\yoruba}{Yor\`ub\'a\xspace}

\newcommand*{\supports}{\texttt{SUPPORTS}}
\newcommand*{\refutes}{\texttt{REFUTES}}
\newcommand*{\NEI}{\texttt{NOT\_ENOUGH\_INFORMATION}}

\newcommand*{\gemma}{\texttt{Gemma}}

\newcommand*{\Qwen}{\texttt{Qwen}}
\newcommand*{\cohere}{\texttt{Cohere}}
\newcommand*{\afriq}{\texttt{AfriqueLLM}}

\newcommand*{\qwenEight}{\texttt{Qwen3-Embedding-8B}}
\newcommand*{\qwenFour}{\texttt{Qwen3-Embedding-4B}}
\newcommand*{\qwenSmall}{\texttt{Qwen3-Embedding-0.6B}}

\newcommand*{\afriEfive}{\texttt{AfriE5-Large-instruct}}

\newcommand*{\mEfiveInstr}{\texttt{multilingual-e5-large-inst.}}
\newcommand*{\mEfive}{\texttt{multilingual-e5-large}}

\newcommand*{\bmTwentyFive}{\texttt{BM25}}

\newcommand*{\afriqgemma}{\texttt{AfriqueGemma-12B}}
\newcommand*{\afriqueqwen}{\texttt{AfriqueQwen-14B}}

\newcommand*{\commandr}{\texttt{Command R v01}}

\newcommand*{\gemmaone}{\texttt{Gemma-3-1B-it}}
\newcommand*{\gemmafour}{\texttt{Gemma-3-4B-it}}
\newcommand*{\gemmatwelve}{\texttt{Gemma-3-12B-it}}
\newcommand*{\gemmatwentyseven}{\texttt{Gemma-3-27B-it}}

\newcommand*{\llamaseventy}{\texttt{Llama-3-70B-instruct}}

\newcommand*{\qwenfourteen}{\texttt{Qwen-3-14B}}

\newcommand*{\tinyaya}{\texttt{Tiny-Aya-Global}}

\newcommand*{\gptfive}{\texttt{GPT-5.2}}

\title{AfrIFact: Cultural Information Retrieval, Evidence Extraction And Fact Checking  for African Languages}

\author{\normalsize 
Israel Abebe Azime$^{1,\dagger}$, Jesujoba O. Alabi$^{1,\dagger}$,
Crystina Zhang$^{2}$,
Iffat Maab$^{3}$, 
 \\ 
\textbf{\normalsize 
Atnafu Lambebo Tonja$^{4,\dagger}$, 
Tadesse Destaw Belay$^{5,\dagger}$,
Folasade Alabi$^{6,\dagger}$,
Salomey osei$^{7,\dagger}$,
} \\
\textbf{\normalsize  
Saminu Muhammad Aliyu$^{8}$,
Nkechinyere Faith Aguobi$^{9,\dagger}$,
Bontu Fufa Balcha$^{11,\dagger}$,
}  \\
\textbf{\normalsize  
Blessing Sibanda$^{\dagger}$,
Davis David$^{10}$,
Mouhamadane Mboup$^{12,\dagger}$,
Daud Abolade$^{\dagger}$,
 }  \\ 
\textbf{\normalsize 
Neo Putini$^{\dagger}$,
Philipp Slusallek$^{1}$, 
David Ifeoluwa Adelani$^{13}$,
Dietrich Klakow$^{1}$   }  \\
\footnotesize
$^{\dagger}$Masakhane NLP, 
$^1$ Saarland University, Saarland Informatic Campus, Germany
$^2$ University of Waterloo, Canada
\\
 \footnotesize
$^3$ National Institute of Informatics, Japan 
$^4$ University College London, UK
$^5$ Instituto Politécnico Nacional, Maxico,
\\ 
 \footnotesize
$^6$ University of Ilorin, Nigeria
$^7$ Universidad de Deusto, Spain,
$^8$ Bayero University,  Nigeria
$^9$ University of Lagos,  Nigeria
\\ 
 \footnotesize
$^{10}$ Black Swan, 
$^{11}$ Addis Ababa University, Ethiopia
$^{12}$ Universite Alioune Diop, Senegal
\\ 
 \footnotesize
$^{13}$ McGill University, Mila-Quebec AI Institute \& Canada CIFAR AI Chair~~~\\
   \small{
   Corresponding author: \href{mailto:se.israel.abebe@gmail.com}{[se.israel.abebe@gmail.com]}
 }
}

\begin{document}
 \maketitle
\begin{abstract}
Assessing the veracity of a claim made online is a complex and important task with real-world implications.  When these claims are directed at communities with limited access to information and the content concerns issues such as healthcare and culture, the consequences intensify, especially in low-resource languages. 
In this work, we introduce \afrifact, a dataset that covers the necessary steps for automatic fact-checking (i.e., information retrieval, evidence extraction, and fact checking),
in ten African languages and English.
Our evaluation results show that even the best embedding models lack cross-lingual retrieval capabilities, and that cultural and news documents are easier to retrieve than healthcare-domain documents, both in large corpora and in single documents. We show that LLMs lack robust multilingual fact-verification capabilities in African languages, while few-shot prompting improves performance by up to 43\% in \afriqueqwen, and task-specific fine-tuning further improves fact-checking accuracy by up to 26\%. These findings, along with our release of the \afrifact~dataset, encourage work on low-resource information retrieval, evidence retrieval, and fact checking.
\end{abstract}

\section{Introduction}
\begin{figure}[!t]
\centering
\includegraphics[width=0.35\textwidth]{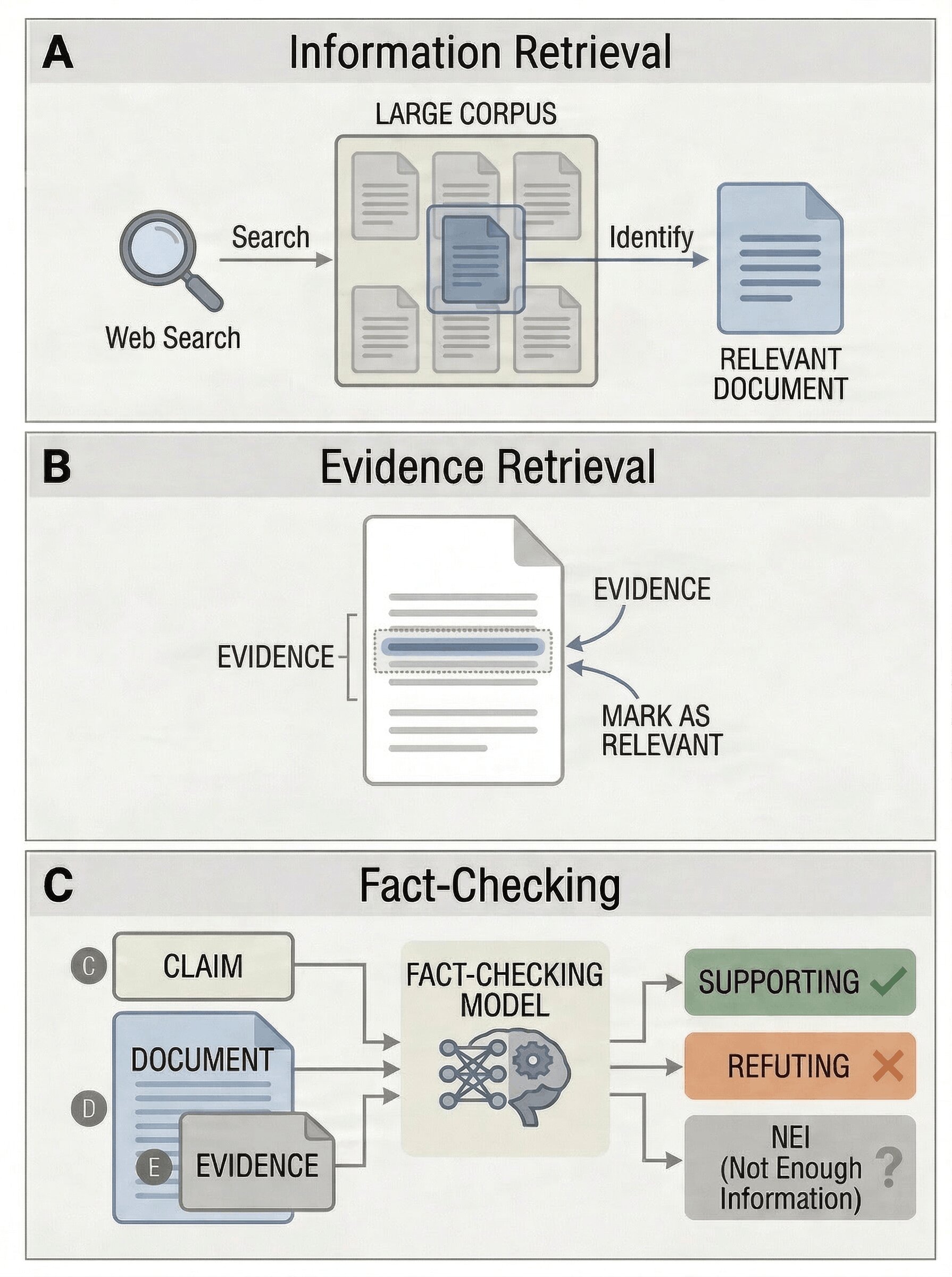}

\caption{
Three steps for verifying factual claims are covered by our dataset.
(A) information retrieval to find documents related to a claim, 
(B) extraction of sentence-level evidence supporting or refuting the claim, and 
(C) fact-checking using the claim, retrieved document, and extracted evidence with LLMs.}
\label{fig:fact-checking}
\end{figure}

\begin{table*}[!ht]
\centering

\resizebox{\textwidth}{!}{
\begin{tabular}{lcccc}
\toprule
\textbf{Dataset} & \textbf{\# Lang.} & \textbf{\# Native African Lang.} & \textbf{Domains} & \textbf{Tasks} \\
\midrule
X-FACT~\cite{gupta-srikumar-2021-x} & 25 & 0 & News, Politics & Claim Verification \\
XFEVER~\cite{chang-etal-2023-xfever} & 6 & 0 & Wikipedia & Multilingual Claim Verification \\
MultiClaim~\cite{panchendrarajan-etal-2025-multiclaimnet} & 78 & 1 (Sudanese) & Social Media, News & Claim Matching, Retrieval \\
Poly-FEVER~\cite{zhang2025poly} & 11 & 1 (Amharic) & Wikipedia, Scientific Articles & Multilingual Fact Verification \\ \midrule
\rowcolor{gray!15}
\textbf{\afrifact (ours)} & \textbf{11} & \textbf{10} & \textbf{Health, Culture(Wiki)-News} & \textbf{Fact Verification, Evidence Extraction, Information Retrieval} \\
\bottomrule
\end{tabular}
}
\caption{\textbf{Comparison of multilingual fact-checking datasets.} We report the number of supported languages, number of African languages, domain coverage, and task types. \textbf{\afrifact} focuses on African languages and introduces multilingual information retrieval, evidence extraction and claim verification across culturally grounded domains.}
\label{tab:fact_datasets}
\end{table*}

Information poverty, defined as the lack of equitable access to information, presents a significant global challenge that disproportionately impacts marginalized communities and regions where low-resource languages are spoken~\cite{GEBREMICHAEL2006267}. With the advent of Large Language Models (LLMs) capable of generating, rewriting, and summarizing information, ensuring the validity and correctness of their outputs remains a critical concern. This issue is especially acute in linguistically diverse contexts, as most current systems are primarily engineered for English-speaking populations~\cite{liu-etal-2025-survey}. Therefore, effectively addressing access to reliable and factual information requires not only improving access but also developing accurate and equitable language technologies, including extracting valuable information from online sources, identifying valid evidence from given documents, and verifying claims with substantiating evidence~\cite{10.1145/3442188.3445922}.

Fact-checking and evidence retrieval~\cite {thorne-etal-2018-fever,schlichtkrull2023averitec} aim to verify the truthfulness of claims using reliable sources. In practice, claims made by individuals or organizations cannot be classified as true or false without supporting evidence. Consequently, real-life fact-checking typically involves three steps: identifying relevant documents or web resources related to a claim, extracting specific sections that support or refute the claim, and classifying the claim as either supported or refuted by the evidence.

In real-world settings, claims disseminated through social media can significantly influence public health decisions, social behaviors, and cultural perspectives. As a result, the ability to reliably evaluate and verify the veracity of such claims is increasingly essential. Furthermore, current systems are rarely evaluated across diverse cultural and linguistic contexts, such as those in Africa, which may undermine reliable access to information by allowing unreliable information to proliferate.


In this work, inspired by the fact-checking process shown in Figure~\ref{tab:fact_datasets}, we introduce \textbf{\afrifact}, a multilingual benchmark containing more than 18,000 claims in 10 African languages and English to assess the capability of  \textbf{embedding models} and \textbf{LLMs} to answer the following research questions: First,  to investigate the capability in extracting relevant documents from large-scale resources, we ask ~\textbf{\textit{(RQ1)}} \textit{What is the effectiveness of embedding models in performing information retrieval for under-resourced African languages?} Secondly, to identify and select the specific sentences or passages that support or refute a claim, assessing the quality and reliability of their evidence extraction, we explore \textbf{\textit{(RQ2)}} \textit{How precisely can embedding models identify and extract relevant evidence from single document for claim verification?} Finally, we ask \textbf{\textit{(RQ3)}} \textit{To what extent can current LLMs correctly verify the factuality of claims?}  by asking LLMs to classify claims as \texttt{SUPPORTED}, \texttt{REFUTED}, or, when the available evidence is insufficient to draw a reliable conclusion, \NEI. To answer this question, this work makes the following contributions:
\begin{itemize}
    \item We introduce \textbf{\afrifact}, a multilingual information retrieval, evidence retrieval and fact checking benchmark covering healthcare, culturally grounded content, with the aim of creating a large-scale evaluation
    ~\footnote{Dataset: \url{https://huggingface.co/collections/masakhane/afrifact} \\ Repository: \url{https://github.com/IsraelAbebe/AfriFact}}. 

    \item We investigate both monolingual and multilingual \textbf{information retrieval} performance using multilingual embedding models.
    
    \item We assess the document-level \textbf{evidence extraction} abilities of embedding models in monolingual, culturally grounded single document settings.

    \item We evaluate the \textbf{fact-checking capabilities} of LLMs across 10 African languages through \textbf{claim classification}, both with and without labeled evidence spans, and examine methods for improvement, such as few-shot prompting and fine-tuning.
\end{itemize}

\section{Related Work}

\paragraph{Automated Fact Checking (AFC)}~ has become an important research area with the goal of combating misinformation by verifying factual claims using computational methods~\cite{vlachos-riedel-2014-fact,thorne-etal-2018-fever,schlichtkrull2023averitec}. Existing research typically model AFC as a multi-stage pipeline that includes claim detection, evidence retrieval, and claim verification~\citep{guo-etal-2022-survey}. Prior work has explored identifying check-worthy claims in several discourse~\citep{Hassan2015DetectingCF}, as well as retrieving evidence from large knowledge sources such as Wikipedia~\citep{thorne-etal-2018-fever,augenstein-etal-2019-multifc} and reasoning over them to verify claims~\citep{nakashole-mitchell-2014-language,thorne-etal-2018-fever,potthast-etal-2018-stylometric}. 

\Cref{tab:fact_datasets} summarizes widely used fact-checking benchmarks, their language coverage, their domains and tasks they can be used for. However, most existing datasets and approaches focus primarily on English, and relatively little work has explored AFC for African languages. MultiClaim~\cite{panchendrarajan-etal-2025-multiclaimnet} is multilingual on a scale, but includes limited African language coverage with one language i.e., primarily Sudanese. In contrast, \afrifact~focuses explicitly on African languages and supports culturally grounded domains, enabling multilingual fact verification with evidence extraction and retrieval.

\paragraph{Information Retrieval (IR)}
Classical evidence retrieval (ER) in AFC courses on retrieving evidence that directly refutes a specific claim~\cite{zheng-etal-2024-evidence}. In contrast, IR is a broader task that aims to retrieve relevant documents for downstream applications, not limited to claim verification. For African languages, several research efforts have focused on information retrieval, including the development of IR datasets such as CLIRMatrix~\citep{sun-duh-2020-clirmatrix}, AfriCLIRMatrix~\citep{ogundepo-etal-2022-africlirmatrix}, MIRACL~\citep{zhang-etal-2023-miracl}, CIRAL~\citep{adeyemi2023ciral}; however, these datasets are often limited in scale or domain coverage, and not designed for fact checking purpose. In this work, we repurpose \afrifact~for IR, enabling benchmarking of embedding models in monolingual and multilingual IR.

\section{\afrifact~Dataset}

\afrifact~covers two domains, health and culture-news, covering ten African languages. These domains were selected because of their high societal impact, their susceptibility to misinformation, and the availability of seed data for dataset curation. 

\textbf{Language Coverage:} Included languages are Amharic, Hausa, Igbo, Oromo, Shona, Swahili, Twi, Wolof, \yoruba, and isiZulu. The selection of these languages was influenced by the number of speakers, the availability of web resources, typological diversity, geographical diversity (West, East and Southern Africa), the coverage in previous datasets, and the availability of annotators. For each selected language, native-speaking annotators were recruited through an African AI community.  
Language coordinators, who are also native speakers, were recruited to oversee the annotators. Linguists were prioritized and, when unavailable, annotators with prior experience in similar projects were prioritized. More details about the languages and annotators are provided in Appendix~\ref{app:data-stats}.

\Cref{fig:dataset-pipeline} shows the data curation pipeline for both domains in \afrifact. In the following sections, we describe the steps in detail.

\begin{figure}[t]
\centering
\includegraphics[width=0.5\textwidth]{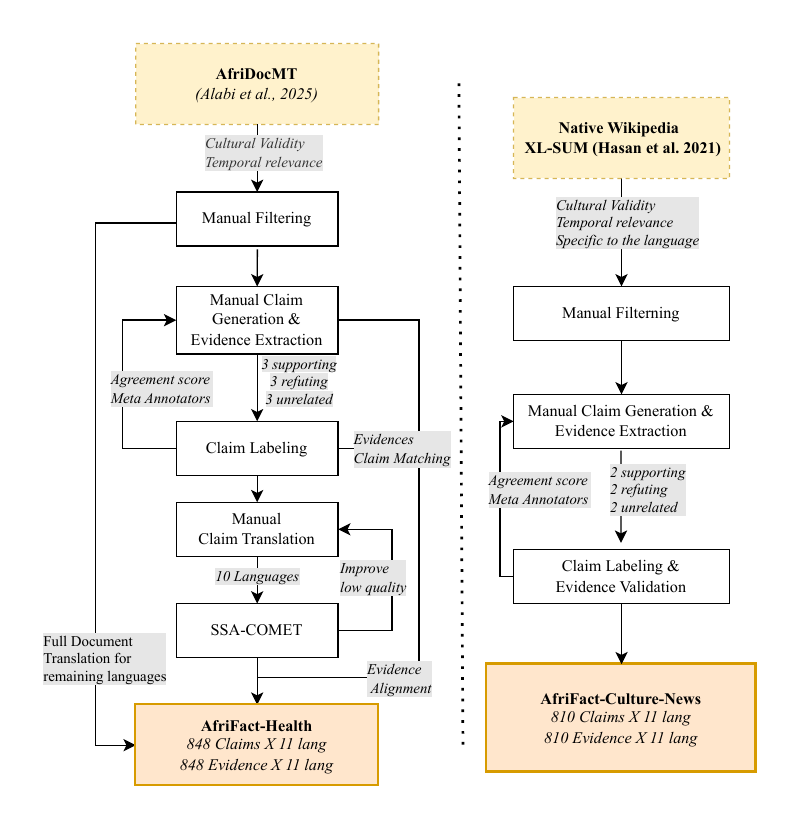}
\caption{ Illustration of the data construction process for \afrifacthealth~on shared health-data approach for lack of structured native health documents and \afrifactculture~culturally grounded, natively sourced data approach 
}
\label{fig:dataset-pipeline}
\end{figure}

\subsection{\afrifacthealth}

To create \afrifacthealth, which focuses on the factuality of healthcare information, particularly common diseases, recurrent outbreaks, and public health concerns, we build on the health split of the AfriDocMT~\citep{alabi-etal-2025-afridoc}. This dataset is a document-level human-translated machine translation (MT) corpus containing 334 articles from the World Health Organization (WHO) website. The articles are translated from English into five African languages, all of which are included in this project. In \afrifacthealth, claim generation, evidence extraction, and factuality labeling are conducted in English. These annotations are then projected to the corresponding African language data through a translation and alignment process.


\textbf{Manual Filtering:} We manually filtered 93 documents from AfriDocMT for use in this work. The selection process was guided by three criteria: cultural relevance and temporal validity. 
To support temporal validity, we excluded time-dependent materials, such as short-term reports, and retained only documents whose content is expected to remain relevant over extended periods. These 93 documents, totaling 2,600 sentences, are translated into five additional African languages not covered by AfriDocMT. 
The translations are done by human expert translators using the same guidelines as in the AfriDocMT dataset.

\textbf{Claim Generation and Evidence Extraction:}
Following the AVeriTeC framework~\citep{schlichtkrull2023averitec}, which guides claim generation via question answering, and considering that \afrifacthealth~ is a multiparallel dataset translated from English, as well as the strong performance of recent LLMs on English tasks, we first generated questions to steer the claim generation process using \texttt{GPT-4.1}. Given the generated questions, annotators, under the supervision of language coordinators, created three supporting, three refuting, and three unrelated (insufficient information) claims per document. The questions helped them focus on important topics and ensure that each claim was relevant and coherent. 
To support or refute claims, annotators also identified evidence spans within the source documents, where evidence is restricted to the sentence level and consists of one or more contiguous full sentences. 
In total, this process produced 848 claims.


\begin{table}[t]
\resizebox{0.45\textwidth}{!}{
    \centering
    \begin{tabular}{l|c|ccc|r}
    \toprule
        \textbf{Split} & \textbf{\# Lang.} & ~ & \textbf{Total Split} & ~ & \textbf{Total} \\ 
        ~ & ~ & \textbf{Train} & \textbf{Val} & \textbf{Test} & ~ \\ \midrule
        \textbf{Health} & 11 & 1100 & 550 & 7678 & ~ \\ 
        \textbf{Culture \& News} & 11 & 1228 & 550 & 7150 & ~ \\ \midrule
        ~ & ~ & 2328 & 1100 & 14828 & 18256 \\ \bottomrule
    \end{tabular}
    }\caption{
\textbf{Dataset statistics} (train, validation, test) for \afrifacthealth~and \afrifactculture~across 11 languages. 
Detail data distribution in Appendix~\ref{app:data-stats}.}
\label{tab:data-stats}    
\end{table}

\textbf{Claim Labeling:} 
Three annotators followed the three-class annotation schema~\cite{thorne-etal-2018-fever} for AFC to label the generated claims, and to verify the associated evidence spans.


\textbf{Manual Claim Translation:} 
At this stage, we have the English documents, generated claims and their labels, corresponding evidence spans. To construct the claims across African languages, we performed manual translation with native speakers for each of the ten target languages. 


\textbf{Translation Quality:} 
To verify the quality of the human translations of documents and claims from English into the ten target languages, we employed SSA-COMET~\cite{li-etal-2025-ssa}. 
Following prior work~\cite{azime2025bridging,alabi-etal-2025-afridoc}, we filtered out translations with scores below 0.6 and engaged language coordinators and translators to either revise them or flag them as low-quality. While SSA-COMET occasionally produced false positives as identified by language coordinators, we found that the metric served as a useful quality-assistance tool for identifying and improving suboptimal translations.

\textbf{Evidence Alignment:} Given the sentence-level parallelism provided within AfriDocMT, we leveraged evidence spans annotated and verified in English to construct multilingual evidence for the \afrifacthealth~ split. This heuristic approach enables a reliable transfer of evidence, as all evidence spans are defined at the sentence level. Subsequently, we involved language coordinators to verify the evidence extracted in each target language and to ensure alignment with the corresponding English evidence.

\subsection{\afrifactculture}
 For \afrifactculture, which focuses on the factuality of information related to cultural knowledge and news, we manually curated Wikipedia articles covering Africa-centric topics inspired by ~\citet{thorne-etal-2018-fever,chang-etal-2023-xfever}. We collected Wikipedia articles in the ten target languages covering locally relevant topics, including political leaders, food, regions, languages, historical events, sports, religion, currencies, and tourism. 
 To further expand regionally relevant content, we added the News domain dataset, which comprises news articles written in the study languages and reporting on regional events. We sourced these articles from XL-Sum~\cite{hasan-etal-2021-xl}, a large-scale news summarization dataset, and retained documents after filtering for document length and safety considerations. 

\textbf{Manual Filtering:} We prioritized culturally focused Wikipedia articles and supplemented them with news documents where cultural coverage was limited, resulting in 200 documents per language, each containing more than 400 sentences and culturally and linguistically authentic documents. 



\textbf{Claim Generation and Evidence Extraction:} To accommodate the diversity of documents, annotators were tasked with generating two \supports, two \refutes, and two \NEI~claims for each document based on selected evidence, each accompanied by annotated evidence sentences. Annotators were also responsible for verifying document quality and validity; if a document contained missing information or did not meet regional relevance criteria, they were instructed to skip it without generating claims.

\textbf{Claim Labeling and Evidence Validation:} Following the same claim labeling procedure used in the health data collection, annotators worked with native-language claims, evidence spans, and source documents to assign of three labels: \supports, \refutes, or \NEI.

\vspace{-1mm}
\paragraph{Agreement Scores}
For both \afrifacthealth~and \afrifactculture~splits, each claim was independently labeled as \supports, \refutes, or \NEI~using three annotators. To assess the reliability of the annotations, we computed Cohen’s kappa \cite{cohen1960coefficient}, Fleiss’ kappa \cite{fleiss1971measuring}, Krippendorff’s alpha \cite{krippendorff2011computing} alpha as inter-annotator agreement measures. As shown in Table \ref{agreement-score}, the dataset achieves substantial agreement across all metrics, indicating high annotation quality and supporting its reliability for evaluating the factuality of large language models.

\begin{table}[t]
    \centering
    \scalebox{0.65}{
    \begin{tabular}{l|l|ccc}
        \hline
        \textbf{Dataset } & \textbf{Metric} & \textbf{REFUTES} & \textbf{NEI} & \textbf{SUPPORTS} \\ \hline
        \textbf{Health} & Fleiss' K. & 0.74 & 0.79 & 0.82 \\ 
        ~ & Cohen's K. & 0.69 & 0.74 & 0.79 \\ 
        ~ & Krippendorff's & 0.74 & 0.79 & 0.82 \\ \hline
        \textbf{Culture-News} & Fleiss' K. & 0.82 & 0.83 & 0.83 \\ 
         & Cohen's K. & 0.81 & 0.82 & 0.82 \\ 
         & Krippendorff's  & 0.85 & 0.85 & 0.85 \\\hline 
    \end{tabular}
    }
    \vspace{-2mm}
    \caption{\textbf{Agreement scores} for English \afrifacthealth~ and averaged per language for \afrifactculture~split. Detailed result found in the Appendix~\ref{app:data-stats}.}
    \label{agreement-score}
\end{table}

\section{Methodology}
Given \afrifact, we address our research questions and detail the experimental setup used in this study.
\paragraph{Information Retrieval Documents:}
\label{sec:dataset:ir_documents}
Our main evaluation setting uses a universal multilingual multi-domain corpus for all types of queries, 
which covers health, news, and Wikipedia articles from all languages.\footnote{
In detail, we include documents from AfriDocMT, XL-Sum news, and manually collected, language-specific Wikipedia articles. 
In addition, we create language-specific retrieval collections using documents exclusively drawn from each corresponding language.
}
This design is to simplify cross-lingual cross-task results comparison on our benchmark.

As data on \afrifacthealth~are parallel across languages, we evaluate it on both monolingual and multilingual settings:\
where the monolingual setting evaluates retrieval results on only the relevant documents from query's language,\footnote{relevant documents that are not in the same language as query are removed from both the retrieval results and labels during evaluation.} and the multilingual setting evaluates retrieval results using relevant documents in all languages.\footnote{Meanwhile, Appendix~\ref{ap:ir_results} provides more evaluation results when retrieving from a corpus containing documents from a single language or a single domain.}


\paragraph{Information Retrieval \& Evidence Extraction Evaluation:}
 In both extraction tasks, we identify the subset of textual units within a document or collection of documents that provide supporting or refuting evidence for a given claim. In our setting, the system retrieves the top-$k$ ranked segments as candidate evidence.  These extracted evidence segments are subsequently used by a downstream fact verification model to determine whether the claim is \supports, \refutes, or \NEI.

For information retrieval following prior literature~\cite{enevoldsen2025mmteb,uemura2025afrimteb}, we use nDCG@10 as the primary metric and also report Recall@100 in the Appendix for reference. Similarly, for Evidence extraction, we reported nDCG@3 as the primary metric and also reported Recall@3 in the Appendix, given the reduced search space for this task.

\paragraph{Claim Labeling Evaluation:} The claim-labeling task asks LLMs to label each document pair and its associated claims as \supports, \refutes, or \NEI. In this work, this evaluation is extended by presenting labeled evidence or balanced few-shot examples. For evidence-based evaluations, we have modified a prompt to guide the LLM to focus on the evidence text extracted from the document, and for few-shot examples, we present three distinct examples with different labels. Evaluation details and reproducibility are explained in Appendix~\ref{reproduce}.

\paragraph{Language Models:}
In this work, we evaluate a diverse set of both embedding-based and lexical retrieval models. For information and evidence extraction, we use models that generate semantically meaningful sentence embeddings. These embeddings can be compared using similarity measures, such as cosine similarity, to rank candidate sentences with respect to a given claim~\cite{reimers-gurevych-2019-sentence}.

\begin{figure*}[t]
    \centering
    \includegraphics[width=\linewidth]{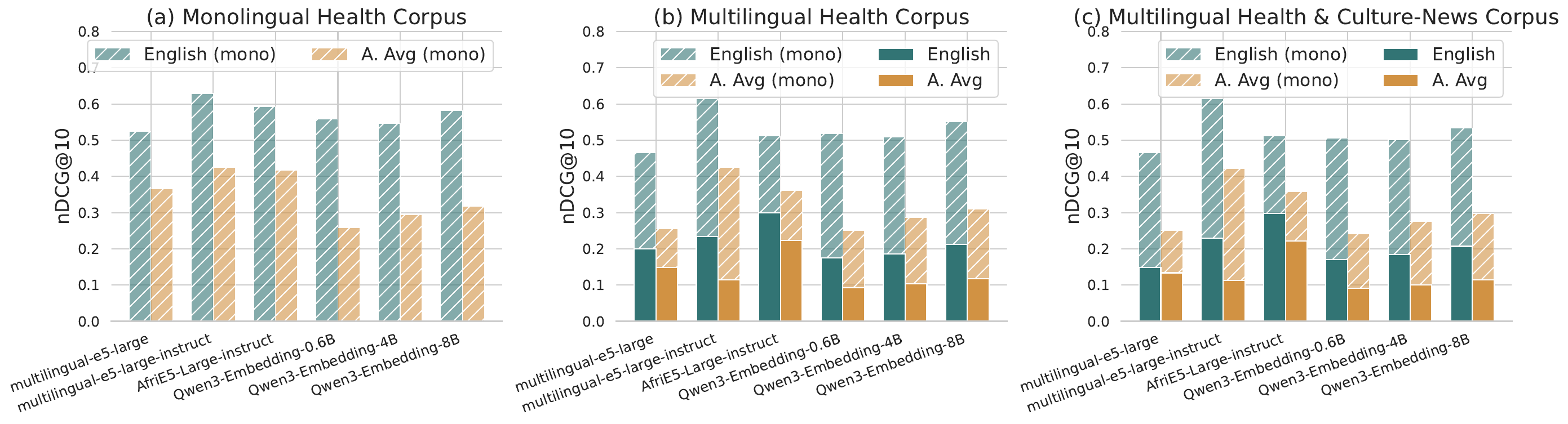}
    \caption{
        nDCG@10 scores on the Health domain when retrieving from different corpora and evaluated on relevant documents in mono- or multi-languages.
        "A. Avg" shows the average score on African languages.
        Scores with label ``mono'' indicate that we only consider relevant documents from the same language as the query during evaluation.
    }
    \label{fig:ir:health}
\end{figure*}

\begin{table*}[!ht]
\centering
\resizebox{0.9\textwidth}{!}{
    \begin{tabular}{l|r|rrrrrrrrrrr|r}
    \toprule
\textbf{model} & \multicolumn{1}{l}{\textbf{English}} & \multicolumn{1}{l}{\textbf{Amharic}} & \multicolumn{1}{l}{\textbf{Hausa}} & \multicolumn{1}{l}{\textbf{Igbo}} & \multicolumn{1}{l}{\textbf{Oromo}} & \multicolumn{1}{l}{\textbf{Shona}} & \multicolumn{1}{l}{\textbf{Swahili}} & \multicolumn{1}{l}{\textbf{Twi}} & \multicolumn{1}{l}{\textbf{Wolof}} & \multicolumn{1}{l}{\textbf{Yoruba}} & \multicolumn{1}{l}{\textbf{Zulu}} & \multicolumn{1}{l}{\textbf{A-Avg.}} & \multicolumn{1}{l}{\textbf{$\Delta$}} \\
\midrule


\multicolumn{13}{c}{\afrifacthealth} \\ 

\mEfive & 0.149 & 0.123 & 0.113 & 0.119 & 0.138 & 0.141 & 0.189 & 0.107 & 0.105 & 0.133 & 0.171 & 0.134 & -0.015 \\
\mEfiveInstr & 0.229 & 0.094 & 0.092 & 0.118 & 0.133 & 0.116 & 0.148 & 0.105 & 0.115 & 0.085 & 0.123 & 0.113 & -0.116 \\
\rowcolor{gray!15}
\afriEfive & 0.298 & 0.235 & 0.221 & 0.248 & 0.215 & 0.224 & 0.287 & 0.183 & 0.173 & 0.172 & 0.257 & 0.222 & -0.077 \\
\qwenSmall & 0.171 & 0.065 & 0.070 & 0.106 & 0.111 & 0.073 & 0.112 & 0.098 & 0.111 & 0.081 & 0.082 & 0.091 & -0.080 \\
\qwenFour & 0.184 & 0.087 & 0.077 & 0.100 & 0.103 & 0.073 & 0.132 & 0.103 & 0.126 & 0.095 & 0.102 & 0.100 & -0.084 \\
\qwenEight & 0.207 & 0.123 & 0.077 & 0.109 & 0.115 & 0.097 & 0.158 & 0.115 & 0.126 & 0.104 & 0.117 & 0.114 & -0.093 \\
\midrule
\multicolumn{13}{c}{\afrifactculture} \\   
\mEfive & 0.597 & 0.487 & 0.341 & 0.471 & 0.261 & 0.394 & 0.406 & 0.286 & 0.303 & 0.173 & 0.329 & 0.345 & -0.252 \\
\rowcolor{gray!15}
\mEfiveInstr & 0.735 & 0.622 & 0.567 & 0.633 & 0.462 & 0.576 & 0.533 & 0.446 & 0.554 & 0.483 & 0.445 & 0.532 & -0.203 \\
\afriEfive & 0.752 & 0.631 & 0.544 & 0.623 & 0.380 & 0.513 & 0.530 & 0.408 & 0.543 & 0.359 & 0.368 & 0.490 & -0.262 \\
\qwenSmall & 0.414 & 0.500 & 0.317 & 0.422 & 0.358 & 0.455 & 0.332 & 0.346 & 0.329 & 0.335 & 0.307 & 0.370 & -0.044 \\
\qwenFour & 0.428 & 0.544 & 0.382 & 0.520 & 0.361 & 0.492 & 0.386 & 0.359 & 0.340 & 0.426 & 0.314 & 0.412 & -0.016 \\
\qwenEight & 0.426 & 0.563 & 0.347 & 0.457 & 0.354 & 0.428 & 0.377 & 0.330 & 0.342 & 0.398 & 0.344 & 0.394 & -0.032 \\
\bottomrule
\end{tabular}
}
\caption{nDCG@10 of Information Retrieval Scores on Multilingual Corpus;
\textbf{A-Avg.} column shows average score on African languages (i.e., excluding English); $\Delta$ column shows the score gap between \textbf{A-Avg.} English scores. gray marks on best A-Avg. scores.}
\label{tab:ir_main}
\end{table*}

For embedding-based retrieval, we evaluate several multilingual and instruction-tuned models, including Qwen3-Embedding models (\texttt{\{0.6B, 4B, 8B\}})~\cite{qwen3embedding}, \mEfive~\cite{wang2024multilingual}, and \afriEfive~\cite{uemura2025afrimteb}.

\begin{table*}[t]
\centering
\resizebox{0.9\textwidth}{!}{
    \begin{tabular}{l|r|rrrrrrrrrrr|r}
    \toprule
\textbf{model} & \multicolumn{1}{l}{\textbf{English}} & \multicolumn{1}{l}{\textbf{Amharic}} & \multicolumn{1}{l}{\textbf{Hausa}} & \multicolumn{1}{l}{\textbf{Igbo}} & \multicolumn{1}{l}{\textbf{Oromo}} & \multicolumn{1}{l}{\textbf{Shona}} & \multicolumn{1}{l}{\textbf{Swahili}} & \multicolumn{1}{l}{\textbf{Twi}} & \multicolumn{1}{l}{\textbf{Wolof}} & \multicolumn{1}{l}{\textbf{Yoruba}} & \multicolumn{1}{l}{\textbf{Zulu}} & \multicolumn{1}{l}{\textbf{A-Avg.}} & \multicolumn{1}{l}{\textbf{$\Delta$}} \\
\midrule
\multicolumn{13}{l}{\afrifacthealth} \\
         \bmTwentyFive & 0.781 & 0.329 & 0.387 & 0.371 & 0.379 & 0.432 & 0.514 & 0.325 & 0.446 & 0.223 & 0.401 & 0.381 & -0.401  \\ 
        \mEfive & 0.770 & 0.363 & 0.415 & 0.349 & 0.386 & 0.450 & 0.541 & 0.333 & 0.418 & 0.288 & 0.438 & 0.398 & -0.372  \\ 
        \mEfiveInstr  & 0.724 & 0.346 & 0.403 & 0.346 & 0.379 & 0.447 & 0.517 & 0.341 & 0.412 & 0.269 & 0.420 & 0.388 & -0.336 \\ 
        \rowcolor{gray!15}
        \afriEfive & 0.742 & 0.371 & 0.419 & 0.361 & 0.395 & 0.455 & 0.527 & 0.363 & 0.443 & 0.290 & 0.442 & 0.407 & -0.335 \\ 
        \qwenSmall & 0.782 & 0.305 & 0.370 & 0.275 & 0.340 & 0.368 & 0.474 & 0.298 & 0.397 & 0.244 & 0.362 & 0.343 & -0.439 \\ 
        \qwenFour  & 0.788 & 0.325 & 0.371 & 0.295 & 0.339 & 0.373 & 0.510 & 0.305 & 0.405 & 0.249 & 0.381 & 0.355 & -0.433 \\ 
        \qwenEight & 0.785 & 0.349 & 0.373 & 0.320 & 0.357 & 0.404 & 0.530 & 0.284 & 0.405 & 0.270 & 0.404 & 0.370 & -0.416  \\  \midrule 
        \multicolumn{13}{l}{\afrifactculture} \\ 
        \rowcolor{gray!15}
        \bmTwentyFive  & 0.751 & 0.628 & 0.555 & 0.646 & 0.520 & 0.796 & 0.617 & 0.483 & 0.447 & 0.346 & 0.382 & 0.542 & -0.209 \\ 
        \mEfive &0.725 & 0.610 & 0.518 & 0.545 & 0.505 & 0.751 & 0.558 & 0.447 & 0.404 & 0.353 & 0.352 & 0.504 & -0.221  \\ 
        \mEfiveInstr & 0.685 & 0.619 & 0.510 & 0.567 & 0.539 & 0.760 & 0.554 & 0.426 & 0.438 & 0.366 & 0.350 & 0.513 & -0.172 \\ 
        \afriEfive & 0.701 & 0.618 & 0.528 & 0.577 & 0.536 & 0.765 & 0.583 & 0.441 & 0.436 & 0.377 & 0.352 & 0.521 & -0.179 \\ 
        \qwenSmall & 0.747 & 0.581 & 0.471 & 0.539 & 0.505 & 0.759 & 0.529 & 0.434 & 0.435 & 0.340 & 0.336 & 0.493 & -0.254  \\ 
        \qwenFour & 0.752 & 0.586 & 0.488 & 0.559 & 0.498 & 0.768 & 0.565 & 0.425 & 0.436 & 0.352 & 0.332 & 0.501 & -0.252 \\ 
        \qwenEight & 0.759 & 0.608 & 0.511 & 0.569 & 0.506 & 0.758 & 0.567 & 0.411 & 0.443 & 0.360 & 0.345 & 0.508 & -0.251 \\ \bottomrule

    \end{tabular}
    }
    \caption{nDCG@3 evidence extraction performances across languages and domains in \afrifact. \textbf{A-Avg.} The column shows the average score on African languages (i.e., excluding English); 
$\Delta$ column shows the score gap between \textbf{A-Avg.} English scores. gray shows the best A-Avg. scores. }
    \label{tab:evidence-extraction-ndcg}
\end{table*}
For the task of claim classification, we evaluate several LLMs from multiple model families, including \gemma~\cite{gemma_2025} 
(\texttt{Gemma-3-{12B \& 27B}-it}), \Qwen~\cite{qwen3technicalreport} (\qwenfourteen), \cohere (\commandr, \tinyaya ~\cite{salamanca2026tiny}), and \afriq~\cite{yu2026afriquellmdatamixingmodel} (\afriqgemma, \afriqueqwen). Almost all models in the main work, with the exception of \tinyaya, have more than 128k context length to eliminate the effect of context size on performance. Additional models are included in the Appendix with lower context sizes. In addition, we include the closed-source model \gptfive~for comparison. Appendix includes more model coverage, while the main paper focuses on a small subset.

\paragraph{Few-shot and Fine-tuning Settings:} For zero-shot settings, we evaluated open LLMs using 3 different prompts, as shown in the appendix, and reported the average scores. For \gptfive, we used a single prompt to reduce cost.

For a few-shot experiment, we selected 3 random shots from the validation set, each with a balanced class label. We limited the number of shots to 3, given the evaluation size and the context length.

For the Finetuning experiment, we created an alpaca-style~\cite{alpaca} version of our training dataset by converting it to an instruction input-output format, and we have both evidence- and no-evidence versions to double our dataset size.  We worked on LoRA~\cite{hu2022lora}, QLoRA~\cite{dettmers2023qlora}, and full fine-tuning settings on \afriqueqwen~for reasons discussed in the results section. Details of reproducibility are explored under Appendix~\ref{reproduce}.

\section{Results and Analysis}

\subsection{Information Retrieval Capability}
\subsubsection{Analysis on \afrifacthealth}

\autoref{fig:ir:health} presents results on \afrifacthealth~queries, under different corpus construction and evaluation settings. 
As \afrifacthealth~provides all-way parallel claims (i.e., query), evidence (i.e., document), and corpus across languages,  it enables a controlled comparison across both query languages 
and target evidence languages.

\paragraph{Query Languages:} 
In \autoref{fig:ir:health}, green bars denote results on English,
whereas orange bars denote the average across the 10 African languages.
Although the subfigures differ in retrieval settings that we will discuss shortly,
a substantial gap between English and African languages is consistently observed across all embedding models.
This suggests considerable room for improving retrieval effectiveness for African languages.

\begin{figure*}[!ht]
    \centering
    \includegraphics[width=0.5\linewidth]{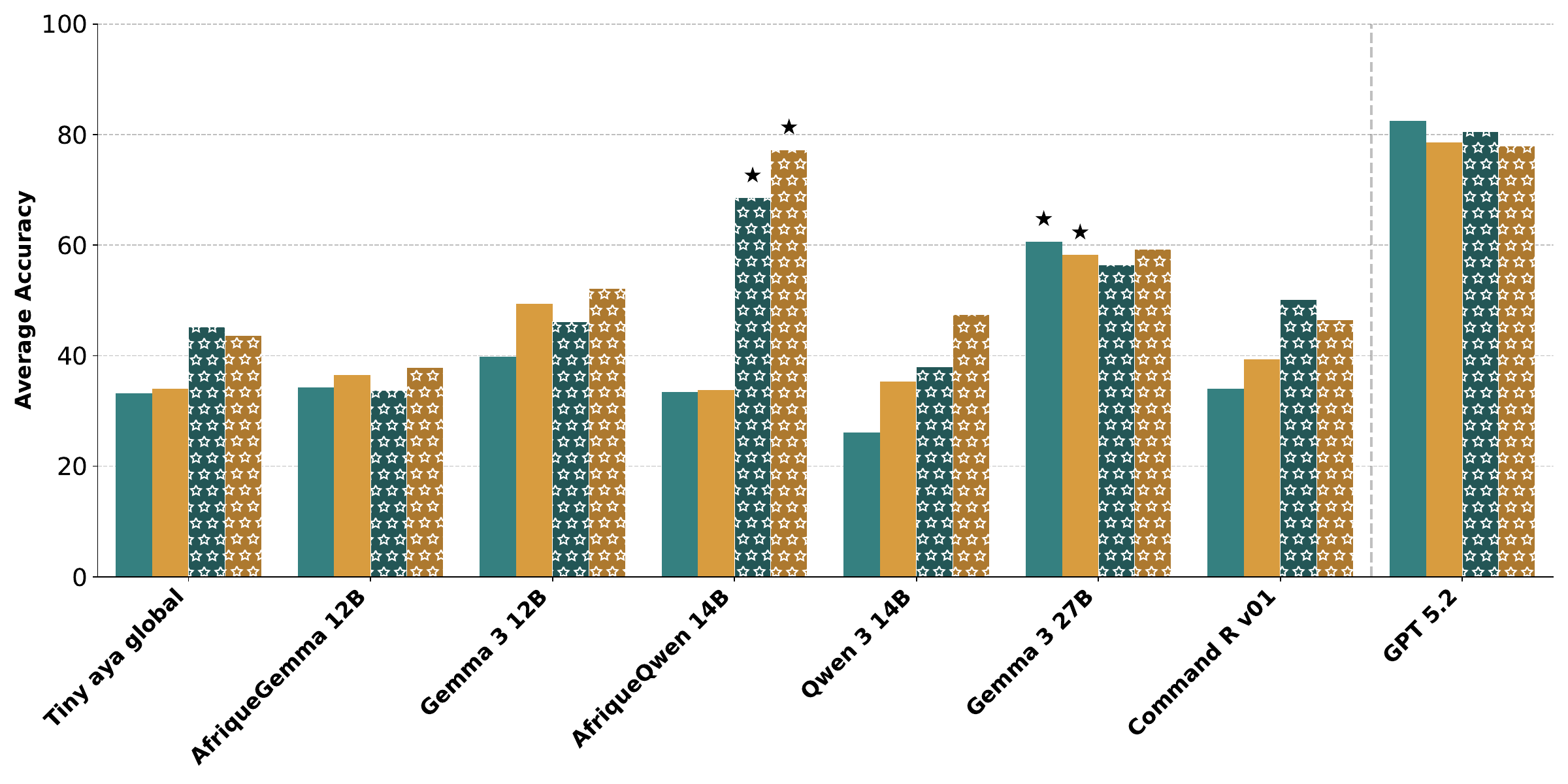}\includegraphics[width=0.5\linewidth]{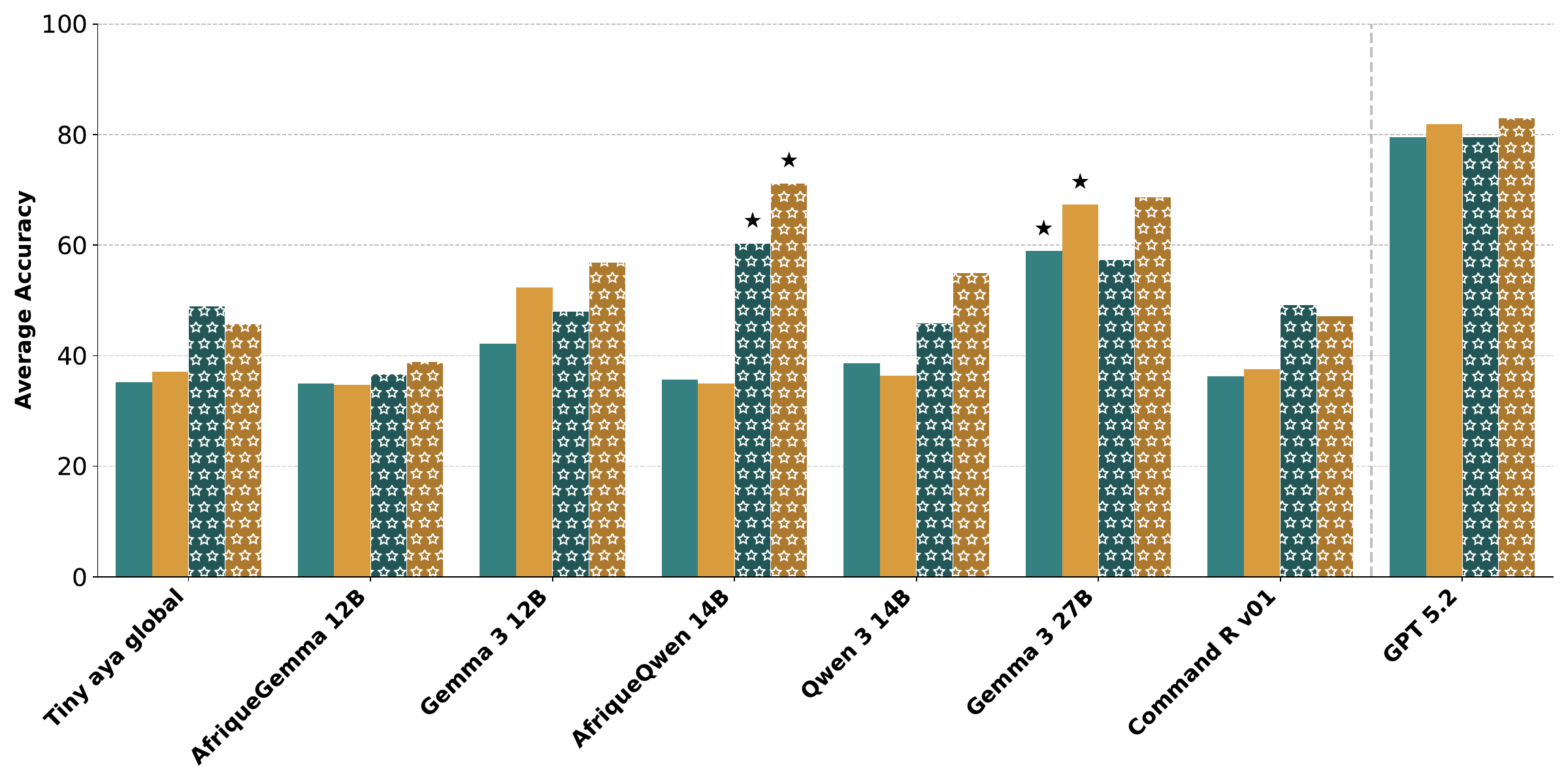}

    \includegraphics[width=0.7\linewidth]{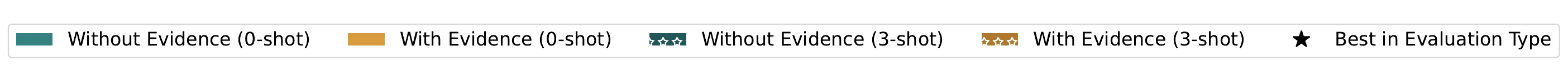}
    \caption{Average accuracy scores on African languages of different language models on the \afrifact~fact-checking task under zero-shot and three-shot settings, with and without evidence. \textbf{Left}: \afrifacthealth~  \textbf{Right}: \afrifactculture}
    \label{fig:claim_classficiation}
\end{figure*}

\paragraph{Document Languages:}
Subfigure (a) serves as the basic retrieval scenario, where each query retrieves from documents in the same language only, i.e., monolingual retrieval; 
In subfigure (b), we expand the corpus to include documents from all languages, resulting in a multilingual retrieval setting. Subfigure (c) further expands the multilingual corpus by adding documents from the Culture-News domain.

As all documents are parallel, each query has relevant documents available in every language when retrieving from the multilingual corpus. On one hand, this helps to evaluate models ``strong-alignment'' capability~\cite{roy-etal-2020-lareqa}, namely, whether it can rank relevant documents in other languages above non-relevant documents in the query language. On the other hand, this is confounded with the impact of corpus size, and how retrieval is affected by enlarging the corpus with non-relevant documents. 

To disentangle these two factors, we report two evaluation variants on subfigures (b) and (c), the solid bars correspond to the standard multilingual setting, which requires retrieving relevant documents from all languages. 
The shaded bars show scores when excluding relevant documents not in the same language as the query (labelled as ``mono''),
which keeps the corpus size almost unchanged but effectively reduces the evaluation criterion back to monolingual evaluation.

Comparing the ``mono'' setting across three subfigures, we observe that increasing corpus size with non-relevant documents indeed brings certain levels of effectiveness degradation, especially moving from subfigure (a) to (b), where the corpus grows by 11 times.
This indicates that embedding models are not particularly robust to a large amount of noisy information.
However, their limitation in retrieving relevant documents across languages is even more severe, as reflected in the large gap between the shaded and solid bars:\ across query languages, performance drops sharply by roughly 50–70\%.

\subsubsection{Main Benchmark}
While the above analysis shows that embedding models are vulnerable in the multilingual retrieval settings, we also find that their relative ranking remains broadly consistent across corpus constructions and evaluation conditions.
Based on this observation, we adopt a unified benchmark corpus that merges documents from all languages and domains, which not only increases task difficulty but also provides a simpler, more holistic evaluation setup.

\autoref{tab:ir_main} reports our main information retrieval results, 
\afrifacthealth~is evaluated on multilingual settings\footnote{corresponding to the non-mono setting in \autoref{fig:ir:health}-(c).},
and \afrifactculture~is evaluated on monolingual settings as it is not parallel across languages.
Comparing the scores, \textbf{we found that embedding models have different behavior across domains.}
where e5-based embeddings (i.e., top-3 rows) are better on \afrifactculture, and Qwen3-based embeddings (bottom-3 rows) are better on \afrifacthealth.


\subsection{Evidence Extraction Capability}

In Table~\ref{tab:evidence-extraction-ndcg}, we present evidence extraction scores and nDCG scores using the top 3 pieces of evidence extracted from a single document presented.
We observe a large gap between English and African languages, except in the case of Shona, for the health domain, with around 0.76 nDCG@3 scores.

\textbf{English African Language evidence retrieval gap is larger on health compared to culture-news.}
The Evidence extraction gap between English and African languages is particularly pronounced in the Health domain compared to the Culture and News domains. In \afrifacthealth, the top-performing English score reaches 0.79, whereas the best-performing African language, Swahili, achieves only 0.52, indicating a substantial disparity. In contrast, the gap narrows in \afrifactculture, where English attains 0.76, and Shona reaches 0.80, even slightly surpassing English. This can be attributed not only to the availability of Wikipedia and cultural domains on the web, but also to a lack of cultural data (i.e., the motivation behind our various dataset-creation strategies). Evidence extraction tasks achieves better performances in \afrifactculture~(0.52-0.56 nDCG@3) compared to results in \afrifacthealth~(0.39-0.44 nDCG@3).

\textbf{Increasing embedding model size is not as important as language coverage.} Despite the substantial parameter differences between models such as \qwenSmall~and \qwenEight, the evidence extraction scores remain very similar. In contrast, \afriEfive{} performs well on \afrifacthealth, while \bmTwentyFive~achieves the highest African-language average on \afrifactculture (0.542), outperforming all embedding models. This highlights the effectiveness of lexical matching for culturally grounded evidence extraction. Performance ranges only from 0.39 to 0.41 in the health domain and from 0.52 to 0.53 in the culture domain. This suggests that scaling model size alone yields limited gains, and that performance differences are more likely driven by dataset characteristics than by model capacity.



\begin{table*}[!ht]
\centering
\resizebox{\textwidth}{!}{
    \begin{tabular}{l|ccccccccccc|r}
    \hline
        \textbf{Model} &   \textbf{Amharic} &  \textbf{English}  & \textbf{Hausa} & \textbf{Igbo} & \textbf{Oromo} & \textbf{Shona} & \textbf{Swahili} & \textbf{Twi} & \textbf{Wolof} & \textbf{Yoruba} &\textbf{Zulu} & \multicolumn{1}{l}{\textbf{A-Avg.}} \\ \midrule
        \multicolumn{13}{l}{\afrifacthealth~ } \\ 
        \textbf{Baseline} & 33.43 / 33.24 &33.43 / 33.24 &33.29 / 34.53 & 33.81 / 33.29  & 33.29 / 33.24 & 33.38 / 33.24  & 33.29 /34.96  & 33.29 / 33.24 &33.33 /33.24  & 33.52 / 33.91 & 33.33 / 34.53 & 33.40 /  33.74  \\
        \textbf{LoRA}  &  49.14 / 63.94 & 49.00 / 59.41 & 47.85 / 61.41 & 46.78 / 56.92 & 48.21 / 58.74 & 48.50 / 58.93 & 48.42 / 62.85 & 37.25 / 40.59 & 38.18 / 42.26 & 46.92 / 59.22 & 48.28 / 60.46  & 45.95 / 56.53 \\
        \rowcolor{gray!15}
        \textbf{QLoRA 8-bit}  & 49.43 / 63.66 & 49.86 / 65.23 & 48.93 / 64.23 & 48.17 / 61.46 & 49.21 / 62.89 & 49.21 / 63.80 & 49.36 / 65.57 & 39.83 / 45.22 & 39.68 / 46.90 & 48.50 / 62.37 & 49.14 / 63.75 &  47.15 / 59.99\\
        \textbf{QLoRA 4-bit}  &39.30 / 50.96 & 34.38 / 35.96 & 36.29 / 49.04 & 35.34 / 44.84 & 36.06 / 44.17 & 36.44 / 44.41 & 37.06 / 49.43 & 33.38 / 34.96 & 33.33 / 34.53 & 38.78 / 49.76 & 37.06 / 50.38  & 36.30 / 45.25 \\
        \textbf{Full Finetuning} & 33.31 / 33.24 & 33.74 / 33.48 & 33.31 / 33.24 & 33.31 / 33.24 & 33.31 / 33.24 & 33.31 / 33.29 & 33.31 / 33.29 & 33.31 / 33.24 & 33.31 / 33.24 & 33.31 / 33.24 & 33.31 / 33.24   & 33.31 / 33.25 \\ \midrule 
        \multicolumn{13}{l}{\afrifactculture~ } \\ 
        \textbf{Baseline} & 35.30/36.10 & 33.40/45.70 & 39.50/36.10 & 33.00 / 33.70 & 34.40 / 35.60 & 38.30 / 37.30 & 36.90 / 32.00 & 31.00 / 31.10 & 38.70 / 37.40 & 32.20 / 33.60 & 38.00 / 34.10 & 35.73 / 34.70 \\ 
        \textbf{LoRA}  & 50.31 / 53.13 & 51.28 / 40.00 & 49.33 / 50.51 & 41.33 / 41.33 & 40.31 / 39.79 & 42.87 / 44.05 & 44.92 / 44.46 & 33.32 / 32.77 & 38.72 / 39.49 & 42.05 / 45.90 & 42.31 / 44.00  & 42.55 / 43.54 \\ 
        \rowcolor{gray!15}
        \textbf{QLoRA 8-bit} & 55.64 / 57.13 & 58.77 / 54.00 & 57.79 / 57.90 & 49.79 / 50.00 & 50.41 / 48.05 & 49.49 / 52.82 & 52.26 / 49.49 & 39.33 / 38.72 & 39.90 / 42.10 & 50.15 / 53.95 & 47.59 / 46.97  & 49.24 / 49.71 \\
        \textbf{QLoRA 4-bit} & 35.18 / 35.62 & 36.05 / 34.97 & 41.49 / 48.82 & 32.87 / 33.13 & 33.44 / 33.49 & 37.95 / 38.36 & 38.46 / 43.08 & 30.82 / 31.08 & 38.56 / 38.62 & 33.59 / 42.05 & 38.00 / 42.46  &  36.04 / 38.67 \\
        \textbf{Full Finetuning} & 33.54 / 33.54 & 33.23 / 33.23 & 40.31 / 40.31 & 31.69 / 31.69 & 32.77 / 32.77 & 37.54 / 37.54 & 36.77 / 36.77 & 30.62 / 30.62 & 38.62 / 38.62 & 32.00 / 32.00 & 37.54 / 37.54  &  35.12 / 35.14 \\\bottomrule
    \end{tabular}
        }
    \caption{\afrifactculture~ and \afrifactculture~without-evidence/with-evidence finetuning experiments on Baseline( \afriqueqwen) finetuned for mixed culture and health data processed for with and without setup. Training approaches include LoRA, QLoRA and full-finetuning }
    \label{tab:finetuning-afriqueqwen}
\end{table*}
\subsection{Fact Checking Capability}
In Figure~\ref{fig:claim_classficiation}, we report African language average scores for models covered. We present $document + claim$, or $document + claim + evidence$ evaluations. Additionally, we show variants of both evaluations by pre-setting 3 balanced few-shot examples aligned with the evaluations. Finally, we include the closed-source model \gptfive~as a closed-source comparison. Details in language scores are shown in Appendix~\ref{sec:detailed-with-without}.

\textbf{Zero-shot fact verification remains extremely challenging across languages.} Mid-size open models show near-random fact checking capability. While larger models like \gemmatwentyseven~score close to 60\% accuracy, most models achieve around 33\% average accuracy on African languages in zero-shot evaluations. Large models such as \gemmatwentyseven and \llamaseventy achieve strong average scores among open-source models. However, improvements from increased model size are often modest compared to gains from few-shot prompting.

\textbf{Few-shot dominates evidence effects.} Across both health and culture-news domains, few-shot examples help improve models\' fact-checking capability. models like \afriqueqwen, \commandr, and \tinyaya~show improved results in few-shot compared to zero-shot. We see significant improvement with few-shot examples, with the biggest being by 43\% from 34\% to 77\% accuracy for \afriqueqwen.

\textbf{Providing evidence does not consistently improve fact-checking accuracy.} Across many models, the difference between with-evidence and without-evidence settings is small or inconsistent, suggesting that models often rely more on their internal knowledge than on explicitly provided evidence.

Looking at Table~\ref{tab:finetuning-afriqueqwen} \afriqueqwen~baseline scores, we see the model has near-random fact-verification performance. This near random performance, which is improved by three shot examples as shown in Appendix~\ref{sec:detailed-with-without}.



\textbf{A small scale dataset, coupled with parameter efficient finetuning, can help improve the performance of models.}~Table~\ref{tab:finetuning-afriqueqwen} shows usage of our multilingual training dataset on a promising model(\afriqueqwen), given its few-shot performance, and we wanted to improve its performance using our small scale dataset. All training configurations except full finetuning seem to boost fact-checking classification. Specifically, QLoRA 8-bit fine-tuning shows the biggest improvement for the \afrifacthealth~domain, with a 26\% average accuracy improvement across African languages. 

\textbf{Low-rank finetuned models learn to use evidence,} shown by all LoRA and QLoRA with evidence models surpassing those without. In our benchmarking experiments, we show that average African language scores with and without evidence are either very similar or that evidence reduces performance, but all LoRA and QLoRA fine-tuning show that models learn to use the provided evidence span and improve scores with evidence present. QLoRA 8-bit quantization shows the best improvement, with evidence of a 12.84\% improvement in the model. Providing evidence substantially improves fact verification in the Health domain but has mixed effects in culture–news, while QLoRA-8 consistently outperforms other adaptation strategies, reaching 58.77\% in the culture-news domain and 65.57\% in the health domain in some languages. Given the consistent performance trends across all languages in model training with a fixed 3-epoch setup and similar hyperparameters, this demonstrates the effectiveness of our multilingual data.



\section{Conclusion and Future Works}

In this work, we introduce \textbf{\afrifact}, a dataset for information retrieval, evidence retrieval, and fact-checking, comprising more than 18,000 human-generated claims, along with annotated source documents and evidence spans across 10 African languages.
Our results show a large gap between information retrieval and evidence extraction scores, even when relevant evidence is found in other languages (i.e., the \afrifacthealth~multilingual experiment), and work done to create Afri-centric models shows promising results. Additionally, we show that open models lack 0-shot fact-verification capabilities, which can be improved with a few-shot examples or dedicated fine-tuning.

We believe this benchmark reveals systematic gaps and challenges in culture-specific information retrieval, evidence retrieval, and fact verification, and encourages researchers to address each of the three problems by expanding dataset coverage and creating robust systems that aim to eliminate misinformation.

\section*{Limitations}
Language inclusion in this dataset was guided by several factors, including the number of speakers, availability of web resources, typological and geographical diversity, coverage in prior datasets, and the availability of annotators.

 \afrifact~focuses on the health and culture-news domains because of their societal importance and exposure to misinformation. However, the need for fact-checking extends far, beyond these areas, expanding AfrIFact to cover other additional domains would offer a more comprehensive assessment of multilingual fact-checking capabilities.

Although the paper discusses a full pipeline for verifying claims by searching documents, extracting evidence spans, and classifying veracity, we treat each step as a standalone section and evaluate it on our dataset. This design allows us to isolate the challenges of each task, which are often studied as separate research problems. Future work can explore end-to-end pipelines that integrate all stages and evaluate their joint performance.

The focus of this work is on creating a substantially larger test set to support robust, high-quality, human-generated, and cross-lingual evaluation, and to accommodate automated fact-checking training datasets. This design choice forced us to create a smaller training and validation set. We intentionally made this choice due to resource constraints and to leave a research gap of fine-tuning better fact verification and extraction models for other works.

Fine-tuning experiments in this work overlook crucial steps such as hyperparameter optimization, data augmentation, and in-depth analysis of trained models, as the scope of this paper is limited to the introduction of dataset creation and benchmarking.

 
\section{Ethical Consideration}

 All annotators, language coordinators, and meta-annotators were compensated fairly in accordance with the agreement(Appendix~\ref{annotators-structure}). Annotators signed contracts specifying the purpose of the data collection and their compensation. We ensured that working conditions were fair and that annotators could skip documents they found unsuitable without penalty. 
 
 The source data for AfrIFact is collected from publicly available resources: AfriDocMT (health documents), Wikipedia (cultural content), and XL-Sum (news articles). No private or personally identifiable information has been collected. The claims were generated by annotators from public data.

\section*{Acknowledgments}
This dataset was created with support from Lacuna Fund and Google.org. The views expressed herein do not necessarily represent those of Lacuna Fund, its Steering Committee, its funders, Meridian Institute or CENIA'

Israel Abebe Azime is funded by the German Federal Ministry of Education and Research and the German federal states (\url{http://www.nhr-verein.de/en/our-partners}) as part of the National High-Performance Computing (NHR) joint funding program.

Also, we would like to express our appreciation to annotators and translators listed below:
\begin{itemize}\itemsep0pt
    \item \textbf{Amharic:} Hana Mekonen, Biniam Asmelash, Bereket Tilahun
    \item \textbf{Twi:} Bernard Opoku, Richmond Opoku, Stephen Arthur
    \item \textbf{Hausa:} Ruqayya Nasir Iro, Maryam Abubakar, Muhammad Abubakar Yaro
    \item \textbf{Igbo:} Kelvin Francis Obitube, Onyinyechi Favour Chibueze, Onyinyechukwu Jane Anowi
    \item \textbf{Oromo:} Dibora Taye, Diribe Kenea, Tadesse Kebede
    \item \textbf{Shona:} Ester Chimhenga, Brian Mupini, Hazel Chamboko
    \item \textbf{Swahili:} Mohamed Mwinyimkuu, Baraka Karuli, Nancy Shao
    \item \textbf{Wolof:} Maimouna Diallo, Sidi Moctar Ndao, Rokhaya Diagne
    \item \textbf{Yoruba:} Simbiat Ajao, Bab\'at\'und\'e P\'op\'o\`o\l{}\'a, Omolade Dorcas
    \item \textbf{isiZulu:} Busisiwe Pakade, Esther Ntuli, Rooweither Mabuya
\end{itemize}



\FloatBarrier
\bibliography{custom}

\clearpage
\appendix

\section{Data Statistics}
\label{app:data-stats}

\afrifact~maintains balanced training and validation sizes per language,  while the test split is substantially larger to support robust, high-quality, human-generated, and cross-lingual evaluation, and to accommodate automated fact-checking training datasets. The languages included are Amharic (Ethiopia), Twi (Ghana), Hausa (Nigeria, Niger, Chad, Cameroon, and Benin), Igbo (Nigeria), Oromo (Ethiopia and Kenya), Shona (Zimbabwe and Mozambique), Swahili (East Africa and the Democratic Republic of Congo), Wolof (Senegal, Gambia, and Mauritania), Yoruba (Nigeria, Benin, and Togo), and isiZulu (South Africa, Lesotho, and Eswatini)

\begin{table*}[!ht]
    \centering
    \scalebox{0.6}{
    \begin{tabular}{l|lllllllllll|l}
    \toprule
        \textbf{language} & \textbf{Amharic} & \textbf{English} & \textbf{Hausa} & \textbf{Igbo} & \textbf{Twi} & \textbf{Oromo} & \textbf{Shona} & \textbf{Swahili} & \textbf{Wolof} & \textbf{Yoruba} & \textbf{Zulu} & \textbf{Total} \\ \midrule
        
         \textbf{\afrifacthealth} & ~ & ~ & ~ & ~ & ~ & ~ & ~ & ~ & ~ & ~ & ~ & ~ \\ 
         SUPPORT & 284 & 284 & 284 & 284 & 284 & 284 & 284 & 284 & 284 & 284 & 284 & 3124 \\ 
        REFUTES & 283 & 283 & 283 & 283 & 283 & 283 & 283 & 283 & 283 & 283 & 283 & 3113 \\ 
        NEI & 281 & 281 & 281 & 281 & 281 & 281 & 281 & 281 & 281 & 281 & 281 & 3091 \\ 
        Total & 848 & 848 & 848 & 848 & 848 & 848 & 848 & 848 & 848 & 848 & 848 & 9328 \\ \midrule
        \textbf{\afrifactculture} & ~ & ~ & ~ & ~ & ~ & ~ & ~ & ~ & ~ & ~ & ~ & ~ \\ 
         SUPPORT & 285 & 263 & 282 & 278 & 278 & 271 & 291 & 307 & 246 & 275 & 286 & 3062 \\ 
        REFUTES & 253 & 267 & 213 & 272 & 272 & 271 & 212 & 211 & 256 & 269 & 225 & 2721 \\ 
        NEI & 275 & 283 & 336 & 263 & 263 & 271 & 310 & 295 & 311 & 269 & 302 & 3178 \\ 
        TOTAL & 810 & 810 & 828 & 810 & 810 & 810 & 810 & 810 & 810 & 810 & 810 & 8928 \\\midrule
        TOTAL & 1658 & 1658 & 1676 & 1658 & 1658 & 1658 & 1658 & 1658 & 1658 & 1658 & 1658 & 18256 \\ \bottomrule
    \end{tabular}
    }\caption{Detailed distribution of annotated examples in the \afrifact~dataset across languages, domains, and verification labels. The table reports the number of instances labeled as \supports, \refutes, and  \NEI~(Not Enough Information) for each language in the \afrifacthealth~and \afrifactculture~domains. }
\end{table*}


\begin{table*}[!ht]
\centering
 \scalebox{0.7}{
    
    \begin{tabular}{l|l|ccc}
        \toprule
        \textbf{Language} & \textbf{Class} & \textbf{Fleiss' Kappa} & \textbf{Cohen's Kappa (avg)} & \textbf{Krippendorff's Alpha} \\ \midrule
        \textbf{Amharic} & \refutes & 0.80 & 0.80 & 0.80 \\ 
        ~ & \NEI & 0.88 & 0.88 & 0.88 \\ 
        ~ & \supports & 0.88 & 0.88 & 0.88 \\ \midrule
        \textbf{Yoruba} & \refutes & 0.92 & 0.92 & 0.92 \\ 
        ~ & \NEI & 0.94 & 0.94 & 0.94 \\ 
        ~ & \supports & 0.96 & 0.96 & 0.96 \\\midrule
        \textbf{igbo} & \refutes & 0.67 & 0.67 & 0.67 \\ 
        ~ & \NEI & 0.76 & 0.76 & 0.76 \\ 
        ~ & \supports & 0.83 & 0.83 & 0.83 \\ \midrule
        \textbf{Swahili} & \refutes & 0.82 & 0.82 & 0.82 \\ 
        ~ & \NEI & 0.82 & 0.83 & 0.83 \\ 
        ~ & \supports & 0.85 & 0.85 & 0.85 \\ \midrule
        \textbf{oromo} & \refutes & 0.87 & 0.87 & 0.87 \\ 
        ~ & \NEI & 0.94 & 0.94 & 0.94 \\ 
        ~ & \supports & 0.87 & 0.87 & 0.87 \\ \midrule
        \textbf{Hausa} & \refutes & 0.74 & 0.75 & 0.75 \\ 
        ~ & \NEI & 0.80 & 0.81 & 0.81 \\ 
        ~ & \supports & 0.86 & 0.87 & 0.87 \\ \midrule
        \textbf{Shona} & \refutes & 0.99 & 0.99 & 0.99 \\ 
        ~ & \NEI & 0.96 & 0.96 & 0.96 \\ 
        ~ & \supports & 0.97 & 0.97 & 0.97 \\ \midrule
        \textbf{Wolof} & \refutes & 0.79 & 0.79 & 0.79 \\ 
        ~ & \NEI & 0.78 & 0.78 & 0.78 \\ 
        ~ & \supports & 0.82 & 0.82 & 0.82 \\\midrule
        \textbf{English} & \refutes & 0.84 & 0.84 & 0.84 \\ 
        ~ & \NEI & 0.72 & 0.72 & 0.72 \\ 
        ~ & \supports & 0.78 & 0.78 & 0.78 \\\midrule
        \textbf{Twi} & \refutes & 0.74 & 0.74 & 0.74 \\ 
        ~ & \NEI & 0.79 & 0.79 & 0.79 \\ 
        ~ & \supports & 0.84 & 0.84 & 0.84 \\\midrule
        \textbf{Zulu} & \refutes & 0.84 & 0.84 & 0.84 \\ 
        ~ & \NEI & 0.72 & 0.72 & 0.72 \\  
        ~ & \supports & 0.77 & 0.78 & 0.77 \\ \bottomrule
    \end{tabular}
    }\caption{Inter-Annotator Agreement Metrics by Language and Category for \afrifactculture, showing substantial to great agreement score.}
\end{table*}

\begin{table*}[!ht]
\centering
\scalebox{0.8}{
\begin{tabular}{lllrr}
\toprule
\textbf{Language}
& \textbf{Family/branch}
& \textbf{Region}
& \textbf{\# speakers}
& \textbf{\# chars in MADLAD (MB)} \\
\midrule
Swahili (swa) & Niger-Congo / Bantu & East \& Central Africa  & 71M--106M  & 2,400MB \\
Hausa (hau) & Afro-Asiatic / Chadic & West Africa    & 77M  & 630MB \\
Amharic (amh)  & Afro-Asiatic / Ethio-Semitic & East Africa    & 57M & 509MB \\
chiShona (sna) & Niger-Congo / Bantu & Southern Africa   & 11M  & 266MB \\
isiZulu (zul) & Niger-Congo / Bantu & Southern Africa  & 27M  & 257MB \\
Igbo (ibo) & Niger-Congo / Volta-Niger & West Africa    & 31M  & 251MB \\
Yoruba (yor) & Niger-Congo / Volta-Niger & West Africa  & 46M & 239MB \\
Oromo (orm) & Afro-Asiatic / Cushitic & East Africa  & 37M & 88MB \\
Twi (twi) & Niger-Congo / Kwa & West Africa  & 9M  & 25MB \\
Wolof (wol) & Niger-Congo / Senegambia & West Africa  & 5M  & 5MB \\
\bottomrule
\end{tabular}
}
\caption{\textbf{Languages represented in \afrifact~dataset.} The table reports the language family and branch, geographic region, countries where the language is widely spoken, and the approximate number of speakers.}
\label{tab:languages}
\end{table*}

\section{Annotators, Structure, Recruitment And Payment}
\label{annotators-structure}
The annotation team consists of 30 annotators, 10 language coordinators, 3 meta-annotators, and 2 administrative assistants. For each language, three native speakers serve as annotators under the supervision of a dedicated language coordinator. The language coordinators have prior experience in data annotation and research and are responsible for guiding the annotation process and ensuring alignment and consistency of the outputs. The recruitment process involved appointing language coordinators based on prior experience in similar projects. Annotators signed contracts specifying the purpose of the data collection and the compensation provided. Everyone involved in the work received training in fact-checking and verification, along with video and textual guidelines for using the tools at each stage of the data annotation process. Before full participation, annotators completed trial tasks that were reviewed and verified by the language coordinators to ensure quality. Appropriate compensation was provided to all contributors involved in the project in accordance with international wage standards.  
\begin{table*}[!ht]
\centering
\resizebox{\textwidth}{!}{
\begin{tabular}{llllc}
\hline
\textbf{Language} & \textbf{Country} & \textbf{Gender Distribution (LC/Annotator)} & \textbf{Qualifications} & \textbf{Resides in Country} \\
\hline
Amharic & Ethiopia & M / 2M1F & Linguistics, CS, Translation & \checkmark \\
Hausa & Nigeria & M / 2M1F & Linguistics and translation & \checkmark \\
Igbo & Nigeria & F / 1M2F & Language studies & \checkmark \\
Twi & Ghana & F / 1M2F & Linguistics, education & \checkmark \\
Oromo & Ethiopia & F/ 2F1M & Linguistics and journalism & \checkmark \\
Shona & Zimbabwe & F / 2M1F & Language studies and translation & \checkmark \\
Swahili & Kenya & M / 2M1F & Linguistics, education & \checkmark \\
Wolof & Senegal & M / 2M1F & Linguistics and translation & \checkmark \\
Yoruba & Nigeria & F / 1M2F & Linguistics, NLP annotation & \checkmark \\
Zulu & South Africa & F / 1M2F & Linguistics and education & \checkmark \\
\hline
\end{tabular}
}
\caption{Summary of annotator teams involved in the creation of the AfriFact dataset. For each language, we report the country context, gender of the language coordinator (LC), gender distribution of the three annotators, a summary of annotator qualifications, and whether annotators reside in the corresponding language region.}
\label{tab:annotators}
\end{table*}
\section{Information Retrieval Results}
\label{ap:ir_results:ndcg_health}
\label{ap:ir_results}
Table \ref{tab:ir_results:ndcg_health} presents additional evaluation results on the \afrifacthealth~dataset as complement to Table~\ref{tab:ir_main}.
Specifically:\ 
\begin{enumerate}[topsep=0pt,itemsep=0pt]
    \item \textit{Monolingual Corpus on Health}:
        Queries from each language only search for relevant documents from documents in the same language under the health domain, e.g., Amharic queries are only searched against Amharic documents. 
    \item \textit{Multilingual Corpus on Health}:
        Queries in each language are searched against documents in all 11 languages. 
        As documents are parallel across languages, now each query has 11 copies of relevant documents, which all contribute to the evaluation of retrieval results. All documents are still from the health domain. 
    \item \textit{Multilingual Corpus on Health and Culture-News}: \textbf{[in main table]}
        Queries in each language are additionally searched against documents in all languages from culture-news domains.
        Evaluations stays the same as \textit{Multilingual Corpus on Health}. 
    \item \textit{
        Multilingual Corpus on Health and Culture-News (eval on monolingual ground-truth
    }: \textbf{[in main table]}
        Identical corpus as above, but exclude relevant documents from non-query-language in retrieval results and ground-truth during evaluation.
\end{enumerate}

In the above settings, the retrieval corpus gradually moved from monolingual mono-domain to multilingual multi-domain, 
where settings 3 and 4 are used as the main evaluation reported in \autoref{tab:ir_main} so as to achieve a unified corpus with culture-news domains. 

Comparing results from the above 4 settings, we have two major observations:\ 
First, \textbf{embedding models struggle with placing relevant documents from non-query languages on top of non-relevant documents from query languages}: Moving from setting 1 to 2, i.e., monolingual-health to multilingual-health, scores of models are at least cut by half if not smaller, and it is not better when using English queries. 

Second, \textbf{embedding models are overall robust to additional non-relevant documents from other languages or domains}. 
Moving from 1 to 4 (corpus includes more non-relevant documents from other languages and domains), while some embeddings are heavily impacted (e.g., \mEfive), the scale is smaller than when introducing relevant documents from other languages as mentioned above. 
Moreover, some embedding models are barely affected by the ``noise'' document --- \mEfiveInstr~maintains the same effectiveness overall on both English and African languages (i.e., 0.629 v.s. 0.614 on English, 0.424 v.s. 0.421 on average on African languages).
The same observation is made moving from 2 to 3 (a multilingual corpus includes more non-relevant documents from other domains), where the overall scores are not impacted.

\begin{table*}[t]

\resizebox{0.98\textwidth}{!}{
    \begin{tabular}{l|r|rrrrrrrrrrr|r}
\toprule
    \textbf{Model} & \textbf{English} & \textbf{Amharic}   & \textbf{Hausa} & \textbf{Igbo} & \textbf{Oromo} & \textbf{Shona} & \textbf{Swahili} & \textbf{Twi} & \textbf{Wolof} & \textbf{Yoruba} &\textbf{Zulu} & \textbf{A-Avg.} & \textbf{$\Delta$}\\ \midrule

\multicolumn{13}{l}{\textit{Monolingual Corpus on Health}} \\
\mEfive & 0.524 & 0.422 & 0.366 & 0.385 & 0.444 & 0.354 & 0.371 & 0.323 & 0.348 & 0.259 & 0.387 & 0.366 & -0.158 \\
\mEfiveInstr & 0.629 & 0.426 & 0.372 & 0.438 & 0.476 & 0.384 & 0.517 & 0.420 & 0.464 & 0.313 & 0.433 & 0.424 & -0.205 \\
\afriEfive & 0.593 & 0.431 & 0.394 & 0.431 & 0.449 & 0.412 & 0.525 & 0.391 & 0.408 & 0.301 & 0.424 & 0.417 & -0.176 \\
\qwenSmall & 0.558 & 0.286 & 0.169 & 0.267 & 0.360 & 0.206 & 0.326 & 0.282 & 0.336 & 0.150 & 0.204 & 0.259 & -0.299 \\
\qwenFour & 0.546 & 0.326 & 0.233 & 0.274 & 0.341 & 0.212 & 0.363 & 0.325 & 0.394 & 0.208 & 0.262 & 0.294 & -0.252 \\
\qwenEight & 0.582 & 0.373 & 0.208 & 0.293 & 0.385 & 0.258 & 0.403 & 0.354 & 0.386 & 0.234 & 0.285 & 0.318 & -0.264 \\

\midrule
\multicolumn{13}{l}{\textit{Multilingual Corpus on Health}} \\
\mEfive & 0.200 & 0.149 & 0.145 & 0.176 & 0.141 & 0.147 & 0.192 & 0.114 & 0.114 & 0.136 & 0.174 & 0.149 & -0.051 \\
\mEfiveInstr & 0.234 & 0.094 & 0.092 & 0.118 & 0.134 & 0.117 & 0.149 & 0.109 & 0.116 & 0.085 & 0.123 & 0.114 & -0.120 \\
\afriEfive & 0.299 & 0.237 & 0.222 & 0.250 & 0.215 & 0.227 & 0.287 & 0.186 & 0.174 & 0.172 & 0.258 & 0.223 & -0.076 \\
\qwenSmall & 0.175 & 0.065 & 0.071 & 0.107 & 0.118 & 0.075 & 0.115 & 0.102 & 0.115 & 0.081 & 0.083 & 0.093 & -0.082 \\
\qwenFour & 0.186 & 0.088 & 0.077 & 0.101 & 0.111 & 0.075 & 0.133 & 0.109 & 0.132 & 0.096 & 0.105 & 0.103 & -0.083 \\
\qwenEight & 0.212 & 0.124 & 0.081 & 0.111 & 0.121 & 0.099 & 0.159 & 0.120 & 0.132 & 0.104 & 0.118 & 0.117 & -0.095 \\

\midrule

\multicolumn{13}{l}{\textit{Multilingual Corpus on Health} (eval on monolingual ground-truth)} \\
\mEfive & 0.448 & 0.367 & 0.208 & 0.165 & 0.377 & 0.150 & 0.180 & 0.226 & 0.215 & 0.117 & 0.205 & 0.221 & -0.227 \\
\mEfiveInstr & 0.613 & 0.426 & 0.372 & 0.436 & 0.472 & 0.384 & 0.517 & 0.417 & 0.461 & 0.313 & 0.428 & 0.423 & -0.190 \\
\afriEfive & 0.501 & 0.385 & 0.295 & 0.270 & 0.335 & 0.311 & 0.376 & 0.326 & 0.335 & 0.236 & 0.316 & 0.319 & -0.183 \\
\qwenSmall & 0.513 & 0.287 & 0.156 & 0.236 & 0.353 & 0.191 & 0.293 & 0.276 & 0.328 & 0.120 & 0.190 & 0.243 & -0.270 \\
\qwenFour & 0.507 & 0.326 & 0.231 & 0.247 & 0.329 & 0.209 & 0.341 & 0.321 & 0.379 & 0.175 & 0.255 & 0.281 & -0.226 \\
\qwenEight & 0.549 & 0.372 & 0.202 & 0.259 & 0.361 & 0.249 & 0.377 & 0.342 & 0.363 & 0.187 & 0.276 & 0.299 & -0.250 \\

\midrule

\multicolumn{13}{l}{\textit{Multilingual Corpus on Health and Culture-News}} \\
\mEfive & 0.149 & 0.123 & 0.113 & 0.119 & 0.138 & 0.141 & 0.189 & 0.107 & 0.105 & 0.133 & 0.171 & 0.134 & -0.015 \\
\mEfiveInstr & 0.229 & 0.094 & 0.092 & 0.118 & 0.133 & 0.116 & 0.148 & 0.105 & 0.115 & 0.085 & 0.123 & 0.113 & -0.116 \\
\afriEfive & 0.298 & 0.235 & 0.221 & 0.248 & 0.215 & 0.224 & 0.287 & 0.183 & 0.173 & 0.172 & 0.257 & 0.222 & -0.077 \\
\qwenSmall & 0.171 & 0.065 & 0.070 & 0.106 & 0.111 & 0.073 & 0.112 & 0.098 & 0.111 & 0.081 & 0.082 & 0.091 & -0.080 \\
\qwenFour & 0.184 & 0.087 & 0.077 & 0.100 & 0.103 & 0.073 & 0.132 & 0.103 & 0.126 & 0.095 & 0.102 & 0.100 & -0.084 \\
\qwenEight & 0.207 & 0.123 & 0.077 & 0.109 & 0.115 & 0.097 & 0.158 & 0.115 & 0.126 & 0.104 & 0.117 & 0.114 & -0.093 \\
\midrule

\multicolumn{13}{l}{\textit{Multilingual Corpus on Health and Culture-News} (eval on monolingual ground-truth)} \\

\mEfive & 0.465 & 0.383 & 0.231 & 0.207 & 0.387 & 0.189 & 0.236 & 0.241 & 0.216 & 0.156 & 0.254 & 0.250 & -0.215 \\
\mEfiveInstr & 0.614 & 0.426 & 0.372 & 0.436 & 0.473 & 0.380 & 0.516 & 0.41 & 0.455 & 0.313 & 0.432 & 0.421 & -0.193 \\
\afriEfive & 0.512 & 0.421 & 0.340 & 0.323 & 0.358 & 0.354 & 0.456 & 0.328 & 0.346 & 0.278 & 0.379 & 0.358 & -0.154 \\
\qwenSmall & 0.506 & 0.284 & 0.161 & 0.250 & 0.324 & 0.195 & 0.308 & 0.259 & 0.314 & 0.128 & 0.194 & 0.242 & -0.264 \\
\qwenFour & 0.501 & 0.323 & 0.229 & 0.251 & 0.299 & 0.209 & 0.353 & 0.303 & 0.362 & 0.181 & 0.250 & 0.276 & -0.225 \\
\qwenEight & 0.534 & 0.368 & 0.192 & 0.272 & 0.341 & 0.246 & 0.387 & 0.333 & 0.343 & 0.216 & 0.278 & 0.298 & -0.236 \\

\bottomrule
\end{tabular}
}

\caption{\textbf{\textit{\afrifacthealth}} nDCG@10 scores when retrieving queries from different corpus and evaluated on corresponding qrels.}
\label{tab:ir_results:ndcg_health}
\end{table*}

\begin{table}[t]
\resizebox{0.95\linewidth}{!}{
    \begin{tabular}{l|c|c|c}
\toprule
    \textbf{Model} & \textbf{English} & \textbf{A-Avg.} & \textbf{$\Delta$}\\ \midrule

\multicolumn{4}{l}{\textit{Monolingual Corpus on Health}} \\
\mEfive & 0.524 & 0.366 & -0.158 \\
\mEfiveInstr & 0.629 & 0.424 & -0.205 \\
\afriEfive & 0.593 & 0.417 & -0.176 \\
\qwenSmall & 0.558 & 0.259 & -0.299 \\
\qwenFour & 0.546 & 0.294 & -0.252 \\
\qwenEight & 0.582 & 0.318 & -0.264 \\

\midrule
\multicolumn{4}{l}{\textit{Multilingual Corpus on Health}} \\
\mEfive & 0.200 & 0.149 & -0.051 \\
\mEfiveInstr & 0.234 & 0.114 & -0.120 \\
\afriEfive & 0.299 & 0.223 & -0.076 \\
\qwenSmall & 0.175 & 0.093 & -0.082 \\
\qwenFour & 0.186 & 0.103 & -0.083 \\
\qwenEight & 0.212 & 0.117 & -0.095 \\


\midrule
\multicolumn{4}{l}{\textit{Multilingual Corpus on Health and Culture-News}} \\
\mEfive & 0.149 & 0.134 & -0.015 \\
\mEfiveInstr & 0.229 & 0.113 & -0.116 \\
\afriEfive & 0.298 & 0.222 & -0.077 \\
\qwenSmall & 0.171 & 0.091 & -0.080 \\
\qwenFour & 0.184 & 0.100 & -0.084 \\
\qwenEight & 0.207 & 0.114 & -0.093 \\

\midrule
\multicolumn{4}{l}{\textit{Multilingual Corpus on Health and Culture-News} (eval on mono)} \\
\mEfive & 0.465 & 0.250 & -0.215 \\
\mEfiveInstr & 0.614 & 0.421 & -0.193 \\
\afriEfive & 0.512 & 0.358 & -0.154 \\
\qwenSmall & 0.506 & 0.242 & -0.264 \\
\qwenFour & 0.501 & 0.276 & -0.225 \\
\qwenEight & 0.534 & 0.298 & -0.236 \\

\bottomrule
\end{tabular}
}

\caption{\textbf{\textit{\afrifacthealth}} nDCG@10 scores when retrieving queries from different corpora and evaluated on corresponding qrels. \autoref{tab:ir_results:ndcg_health} reports results per-language.}
\label{tab:ir_results:ndcg_health_compact}
\end{table}

\section{Evidence Extraction Detailed Metrics}
\label{app:evidence-retrival}
Table~\ref{tab:recall-evidence-extraction} shows recall@3 alternative metric scores for evidence retrieval from a single document.
\begin{table*}[!ht]
\caption{\afrifact~Recall@3 Evidence Extraction Scores.}
\centering
\resizebox{\textwidth}{!}{
    \begin{tabular}{l|r|rrrrrrrrrrr|r}
    \toprule
\textbf{model} & \multicolumn{1}{l}{\textbf{English}} & \multicolumn{1}{l}{\textbf{Amharic}} & \multicolumn{1}{l}{\textbf{Hausa}} & \multicolumn{1}{l}{\textbf{Igbo}} & \multicolumn{1}{l}{\textbf{Oromo}} & \multicolumn{1}{l}{\textbf{Shona}} & \multicolumn{1}{l}{\textbf{Swahili}} & \multicolumn{1}{l}{\textbf{Twi}} & \multicolumn{1}{l}{\textbf{Wolof}} & \multicolumn{1}{l}{\textbf{Yoruba}} & \multicolumn{1}{l}{\textbf{Zulu}} & \multicolumn{1}{l}{\textbf{A-Avg.}} & \multicolumn{1}{l}{\textbf{$\Delta$}} \\ \midrule
        \multicolumn{13}{l}{\afrifacthealth} \\    
        \mEfive & 0.820 & 0.408 & 0.460 & 0.403 & 0.439 & 0.509 & 0.586 & 0.396 & 0.484 & 0.325 & 0.495 & 0.450 & -0.369 \\ 
        \mEfiveInstr & 0.822 & 0.380 & 0.455 & 0.401 & 0.422 & 0.507 & 0.572 & 0.398 & 0.473 & 0.317 & 0.478 & 0.440 & -0.382 \\ 
        \rowcolor{gray!15}
       \afriEfive & 0.822 & 0.412 & 0.464 & 0.412 & 0.437 & 0.513 & 0.574 & 0.409 & 0.506 & 0.337 & 0.495 & 0.456 & -0.366  \\ 
       \qwenSmall & 0.823 & 0.355 & 0.444 & 0.344 & 0.402 & 0.444 & 0.548 & 0.350 & 0.470 & 0.300 & 0.417 & 0.407 & -0.416 \\ 
        \qwenFour & 0.825 & 0.373 & 0.443 & 0.349 & 0.395 & 0.458 & 0.567 & 0.347 & 0.480 & 0.294 & 0.440 & 0.415 & -0.410 \\ 
        \qwenEight & 0.827 & 0.377 & 0.424 & 0.381 & 0.408 & 0.478 & 0.587 & 0.344 & 0.473 & 0.317 & 0.462 & 0.425 & -0.402\\ \midrule
        \multicolumn{13}{l}{\afrifactculture} \\  
        \mEfive  & 0.807 & 0.637 & 0.589 & 0.606 & 0.560 & 0.808 & 0.616 & 0.512 & 0.474 & 0.391 & 0.390 & 0.558 & -0.248 \\ 
        \mEfiveInstr & 0.790 & 0.644 & 0.572 & 0.604 & 0.595 & 0.805 & 0.616 & 0.491 & 0.489 & 0.419 & 0.394 & 0.563 & -0.227  \\ 
        \rowcolor{gray!15}
        \afriEfive & 0.797 & 0.639 & 0.585 & 0.632 & 0.595 & 0.808 & 0.638 & 0.498 & 0.481 & 0.423 & 0.395 & 0.569 & -0.228 \\ 
        \qwenSmall & 0.825 & 0.613 & 0.552 & 0.591 & 0.570 & 0.801 & 0.603 & 0.495 & 0.484 & 0.391 & 0.379 & 0.548 & -0.277 \\ 
        \qwenFour  & 0.848 & 0.623 & 0.572 & 0.618 & 0.573 & 0.810 & 0.638 & 0.483 & 0.479 & 0.396 & 0.366 & 0.556 & -0.292 \\ 
        \qwenEight  & 0.836 & 0.634 & 0.581 & 0.627 & 0.571 & 0.803 & 0.638 & 0.473 & 0.484 & 0.410 & 0.382 & 0.560 & -0.276 \\ \bottomrule
    \end{tabular}
    }
    \label{tab:recall-evidence-extraction}
\end{table*}

\section{Fact Classification Detailed Scores}
\label{sec:detailed-with-without}

In Table \ref{tab:with-without-detailed}, we observe that providing labeled evidence does not consistently improve fact-checking accuracy. Many models collapse to near-random performance in the zero-shot setting, indicating difficulty in understanding the task without demonstrations. Few-shot examples often yield dramatic improvements of up to 43 points for models such as \afriqueqwen. Increasing model size generally improves performance, but gains plateau beyond mid-scale models. English consistently achieves higher scores than African languages, highlighting a persistent multilingual performance gap. Languages such as Wolof, Twi, and Yoruba show lower scores across most models, while \gptfive~demonstrates substantially higher performance across all evaluation settings.

{\footnotesize
\begin{table*}[!ht]
\centering
\caption{\textbf{\afrifacthealth}~ 0-shot/3-shot Accuracy Scores. The best open source models are Bold, and the best overall models are grayed. }
\resizebox{\textwidth}{!}{
    \begin{tabular}{l|ccccccccccc|r}
    \toprule
        \textbf{Model} & \textbf{Amharic} & \textbf{English} & \textbf{Hausa} & \textbf{Igbo} & \textbf{Oromo} & \textbf{Shona} & \textbf{Swahili} & \textbf{Twi} & \textbf{Wolof} & \textbf{Yoruba} &\textbf{Zulu} & \textbf{A-Avg.} \\ 
        \midrule
        \multicolumn{13}{l}{\afrifacthealth~ With Evidence} \\ 
        \afriqgemma & 38.11/38.68 & 36.01/47.66 & 36.82/38.44 & 38.59/40.40 & 38.68/39.40 & 35.20/41.17 & 34.48/38.40 & 34.81/34.48 & 34.19/33.38 & 35.96/35.86 & 37.73/38.78 & 36.46/37.90 \\
        \afriqueqwen & 33.24/71.11 & 33.24/92.26 & 34.53/84.19 & 33.29/83.38 & 33.24/86.44 & 33.24/86.87 & 34.96/86.34 & 33.24/58.17 & 33.24/54.97 & 33.91/79.27 & 34.53/81.76 & 33.74/\textbf{77.25} \\
        \commandr & 39.06/35.24 & 63.75/89.83 & 40.54/45.08 & 38.49/44.56 & 36.91/55.06 & 39.54/47.66 & 45.32/46.94 & 36.68/48.95 & 35.77/52.96 & 37.73/38.97 & 42.88/49.28 & 39.29/46.47 \\
        \gemmatwelve & 38.59/39.02 & 58.07/71.30 & 55.21/58.31 & 58.60/60.74 & 52.58/53.39 & 56.26/63.18 & 58.36/63.80 & 45.46/47.23 & 38.40/43.36 & 35.77/35.91 & 54.58/56.30 & 49.38/52.13\\
        \gemmaone & 34.77/35.91 & 34.34/52.39 & 34.24/35.51 & 33.76/36.51 & 33.67/33.72 & 33.76/37.18 & 35.82/39.21 & 33.67/35.10 & 33.43/34.15 & 34.62/34.03 & 34.00/35.24 & 34.17/35.65  \\
        \gemmatwentyseven  & 40.16/40.15 & 84.38/88.97 & 64.33/69.70 & 68.10/68.02 & 64.47/65.83 & 73.93/75.82 & 72.87/78.40 & 47.18/45.95 & 47.85/43.70 & 38.40/38.11 & 65.09/67.19 & \textbf{58.24}/59.29 \\
        \gemmafour & 35.43/35.96 & 58.40/74.93 & 44.94/47.99 & 46.08/46.18 & 44.70/41.45 & 55.92/54.35 & 54.63/62.89 & 40.83/37.92 & 41.31/37.01 & 34.72/34.72 & 48.90/50.00 & 44.75/44.85 \\
        \llamaseventy & 49.90/42.17 & 79.13/89.68 & 60.70/73.97 & 63.51/73.69 & 61.27/69.58 & 56.06/65.52 & 67.72/86.44 & 57.55/59.22 & 53.77/60.46 & 50.48/43.74 & 57.50/67.14 & 57.85/64.19 \\
        \qwenfourteen & 36.29/63.56 & 34.67/73.59 & 33.52/45.56 & 33.81/43.70 & 35.82/40.69 & 34.57/44.56 & 39.97/55.44 & 33.24/45.89 & 35.48/41.98 & 34.48/42.31 & 35.82/51.00 & 35.30/47.47 \\
        \tinyaya  & 33.24/44.70 & 33.38/68.15 & 34.05/42.36 & 33.24/47.80 & 33.24/36.96 & 33.29/51.43 & 37.63/48.90 & 33.24/34.91 & 33.24/45.80 & 34.19/36.87 & 34.72/46.56 & 34.01/43.63 \\
        \rowcolor{gray!15}
        \gptfive & 80.95/78.94 & 83.24/84.10 & 78.22/77.65 & 79.51/79.94 & 78.65/78.65 & 80.80/79.80 & 82.52/80.23 & 75.64/76.36 & 73.93/74.50 & 78.51/77.08 & 77.36/75.79 & 78.61/77.89 \\\midrule
        \multicolumn{13}{l}{\afrifacthealth~ Without Evidence} \\ 
        \afriqgemma & 34.86/33.62 & 37.01/53.49 & 34.34/35.77 & 34.15/34.48 & 34.29/32.38 & 34.81/39.88 & 34.48/38.59 & 34.10/29.51 & 33.86/31.81 & 33.24/28.75 & 34.57/32.38 & 34.27/33.72  \\
        \afriqueqwen & 33.43/73.73 & 33.43/79.42 & 33.29/72.54 & 33.81/71.30 & 33.29/72.87 & 33.38/72.45 & 33.29/75.64 & 33.29/54.63 & 33.33/49.43 & 33.52/70.92 & 33.33/71.87 & 33.40/\textbf{68.54} \\
        \commandr & 34.86/39.45 & 51.05/80.04 & 33.67/47.66 & 33.81/47.33 & 33.48/56.69 & 33.67/51.00 & 36.58/53.34 & 33.57/50.00 & 33.43/52.01 & 33.43/46.13 & 34.19/57.93 & 34.07/50.15 \\
        \gemmatwelve & 50.91/55.01 & 58.55/70.15 & 39.92/53.34 & 39.40/43.31 & 38.44/43.98 & 40.74/52.87 & 47.99/59.84 & 33.14/35.48 & 32.81/35.10 & 31.76/33.24 & 42.55/48.95 & 39.77/46.11 \\
        \gemmaone & 34.81/36.53 & 37.01/51.58 & 33.43/35.74 & 33.57/34.89 & 34.15/33.98 & 35.48/37.42 & 35.05/41.79 & 33.29/34.36 & 35.00/33.93 & 33.24/33.21 & 34.10/34.17 & 34.21/35.60 \\
        \gemmatwentyseven  & 71.74/67.87 & 79.51/76.32 & 67.80/64.53 & 63.68/59.60 & 59.74/55.65 & 70.24/66.15 & 74.36/71.61 & 44.23/38.64 & 44.59/37.30 & 45.24/43.67 & 64.90/58.68 & \textbf{60.65}/56.37  \\
        \gemmafour & 33.14/53.10 & 41.26/67.81 & 30.85/49.81 & 30.95/39.40 & 28.46/39.59 & 29.70/47.85 & 33.91/58.36 & 27.84/35.58 & 25.26/40.35 & 30.52/32.62 & 33.38/46.56 & 30.40/44.32 \\
        \llamaseventy & 51.09/50.86 & 70.95/79.37 & 56.50/68.39 & 55.53/64.66 & 47.68/60.08 & 48.51/57.69 & 61.72/75.17 & 47.74/51.58 & 42.75/52.63 & 45.07/47.18 & 54.50/59.89 & 51.11/58.81 \\
        \qwenfourteen & 29.99/34.19 & 28.61/60.98 & 25.74/39.68 & 24.21/37.15 & 28.13/37.01 & 25.07/38.63 & 27.79/40.74 & 24.83/36.25 & 27.08/39.06 & 25.93/35.77 & 22.78/40.83 & 26.16/37.93 \\
        \tinyaya & 33.29/50.96 & 33.24/62.23 & 33.24/47.23 & 33.19/46.90 & 33.29/37.82 & 33.29/49.71 & 33.29/53.15 & 33.29/34.96 & 33.29/41.55 & 33.05/43.51 & 33.19/46.47 & 33.24/45.22 \\
        \rowcolor{gray!15}
        \gptfive & 86.96/83.38 & 88.25/86.82 & 82.95/80.37 & 83.67/81.38 & 82.52/80.95 & 85.10/81.66 & 84.81/82.23 & 79.23/78.94 & 77.36/78.08 & 82.09/79.37 & 80.66/79.08 & 82.54/80.54 \\ \bottomrule
    \end{tabular}
        }
    \label{tab:with-without-detailed}
\end{table*}
}

{\footnotesize
\begin{table*}[!ht]
\centering
\caption{\textbf{\afrifactculture}~ 0-shot/3-shot Accuracy Scores. The best open source models are Bold, and the best overall models are grayed.}
\resizebox{\textwidth}{!}{
    \begin{tabular}{l|ccccccccccc|r}
    \toprule
        \textbf{Model} &   \textbf{Amharic} &  \textbf{English}  & \textbf{Hausa} & \textbf{Igbo} & \textbf{Oromo} & \textbf{Shona} & \textbf{Swahili} & \textbf{Twi} & \textbf{Wolof} & \textbf{Yoruba} &\textbf{Zulu} & \textbf{A-Avg.} \\
        \midrule
        \multicolumn{13}{l}{\afrifactculture~ With Evidence} \\ 
        \afriqgemma & 36.10/39.10 & 45.70/57.70 & 36.10/44.50 & 33.70/36.70 & 35.60/39.00 & 37.30/48.20 & 32.00/34.20 & 31.10/32.30 & 37.40/36.70 & 33.60/35.50 & 34.10/43.40 & 34.70/38.96 \\ 
        \afriqueqwen & 33.50/77.00 & 35.40/99.90 & 37.60/77.10 & 31.80/78.40 & 32.80/80.80 & 37.50/76.80 & 36.60/71.80 & 30.60/61.10 & 38.60/44.70 & 34.00/74.60 & 36.80/70.10 & 34.98/\textbf{71.24} \\ 
        \commandr & 42.80/39.20 & 53.60/92.80 & 37.50/47.10 & 32.60/46.30 & 33.00/57.50 & 38.70/43.50 & 39.10/58.20 & 31.30/48.90 & 38.60/42.60 & 37.60/41.70 & 43.80/47.10 & 37.50/47.21 \\ 
        \gemmatwelve & 61.60/65.10 & 82.60/86.40 & 59.50/63.10 & 58.10/60.70 & 47.60/53.60 & 51.00/63.50 & 56.40/57.80 & 45.60/50.70 & 39.40/44.50 & 49.80/52.70 & 54.40/57.00 & 52.34/56.87 \\ 
        \gemmaone & 35.00/33.70 & 35.20/54.70 & 38.10/37.00 & 31.80/38.70 & 32.80/38.00 & 37.70/33.60 & 36.30/30.80 & 30.70/33.80 & 38.70/37.20 & 35.10/36.90 & 37.50/32.70 & 35.37/35.24 \\ 
        \gemmatwentyseven & 80.60/81.80 & 98.70/98.80 & 70.90/73.60 & 73.90/75.30 & 69.20/68.00 & 74.10/76.60 & 65.30/66.10 & 59.80/61.30 & 48.40/49.00 & 64.20/64.90 & 67.40/70.10 & \textbf{67.38}/68.67 \\ 
        \gemmafour & 58.70/61.30 & 63.00/75.40 & 51.70/56.30 & 49.80/51.00 & 46.50/49.20 & 50.90/54.90 & 52.60/55.10 & 46.20/45.10 & 40.40/40.60 & 44.40/46.80 & 46.60/52.30 & 48.78/51.26 \\ 
        \qwenfourteen & 38.50/67.30 & 46.80/72.10 & 35.90/54.60 & 37.10/53.20 & 36.60/49.70 & 37.30/64.60 & 37.70/50.50 & 36.60/53.80 & 31.60/49.10 & 35.50/51.00 & 37.30/55.70 & 36.41/54.95 \\ 
        \tinyaya & 33.50/57.30 & 40.60/64.20 & 43.90/41.10 & 32.10/50.50 & 32.80/41.60 & 37.50/52.10 & 40.80/46.80 & 30.60/43.50 & 38.60/41.30 & 39.20/39.30 & 42.30/43.70 & 37.13/45.72 \\ 
        \rowcolor{gray!15}
        \gptfive & 89.69/89.23 & 97.54/99.38 & 86.46/85.85 & 84.15/87.23 & 80.15/81.54 & 85.38/82.62 & 71.23/73.38 & 78.00/81.38 & 79.23/79.54 & 82.77/86.00 & 82.00/83.54 & 81.91/83.03 \\ \midrule
        \multicolumn{13}{l}{\afrifactculture~ Without Evidence} \\ 
        \afriqgemma & 33.60/34.70 & 33.70/34.10 & 40.10/40.90 & 31.30/32.70 & 32.90/33.70 & 37.40/45.10 & 36.10/37.70 & 30.70/32.80 & 37.60/36.80 & 32.20/34.10 & 37.50/38.10 & 34.94/36.66 \\ 
        \afriqueqwen & 35.30/66.30 & 33.40/86.90 & 39.50/66.50 & 33.00/67.30 & 34.40/64.80 & 38.30/67.50 & 36.90/57.90 & 31.00/52.70 & 38.70/38.20 & 32.20/59.50 & 38.00/62.80 & 35.73/\textbf{60.35} \\ 
        \commandr & 38.30/41.60 & 43.60/80.70 & 39.90/47.70 & 32.60/52.50 & 33.40/57.80 & 38.60/49.50 & 38.60/54.90 & 31.30/50.90 & 38.90/39.80 & 32.40/48.10 & 38.80/49.50 & 36.28/49.23 \\ 
        \gemmatwelve & 41.20/45.40 & 45.00/51.10 & 49.50/55.80 & 39.40/43.40 & 38.50/46.00 & 47.50/57.90 & 46.90/52.40 & 36.60/41.50 & 38.60/40.50 & 36.30/42.30 & 47.80/55.20 & 42.23/48.04 \\ 
        \gemmaone & 34.30/35.60 & 33.60/43.80 & 40.20/35.00 & 31.70/36.70 & 32.60/37.50 & 37.50/38.10 & 36.90/30.40 & 30.80/34.30 & 38.70/35.30 & 32.70/37.50 & 37.60/34.40 & 35.30/35.48 \\ 
        \gemmatwentyseven & 59.90/60.70 & 67.10/65.90 & 68.40/65.60 & 58.70/56.70 & 60.10/56.90 & 66.30/65.20 & 64.80/63.40 & 51.00/50.60 & 43.80/39.60 & 54.00/52.70 & 62.50/62.90 & \textbf{58.95}/57.43 \\ 
        \gemmafour & 38.50/42.50 & 42.50/48.60 & 47.30/53.30 & 36.50/40.50 & 35.10/40.30 & 41.10/49.70 & 43.80/51.50 & 34.50/37.70 & 38.50/39.60 & 34.50/41.50 & 41.70/49.60 & 39.15/44.62 \\ 
        \qwenfourteen & 37.60/41.60 & 47.50/54.40 & 42.20/48.80 & 38.50/46.20 & 38.20/46.60 & 39.40/53.10 & 44.10/43.30 & 36.70/48.30 & 33.70/40.30 & 38.10/43.60 & 38.10/47.50 & 38.66/45.93 \\ 
        \tinyaya & 33.50/55.40 & 33.50/50.50 & 40.40/47.40 & 31.70/51.60 & 32.80/45.90 & 37.50/53.90 & 36.80/52.50 & 30.60/45.20 & 38.60/41.30 & 32.00/48.20 & 37.50/48.50 & 35.14/48.99 \\
        \rowcolor{gray!15}
        \gptfive & 84.00/83.23 & 97.54/98.00 & 81.54/83.08 & 84.92/85.54 & 87.23/87.54 & 74.31/74.31 & 71.23/72.00 & 91.54/89.54 & 70.62/70.46 & 81.08/81.08 & 69.23/68.77 & 79.57/79.55 \\ \bottomrule
        
    \end{tabular}
        }
    \label{tab:with-without-detailed}
\end{table*}

\begin{table*}[t]
\centering
\resizebox{\textwidth}{!}{%
\begin{tabular}{lcccccccccc}
\toprule
Model & Amharic & Hausa & Igbo & Oromo & Shona & Swahili & Twi & Wolof & Yoruba & Zulu \\
\midrule

\multicolumn{11}{l}{\textbf{AfrIFact-Health -- With Evidence (0-shot)}} \\
AfriqueQwen-14B
& 0.332 $\pm$ 0.000 & 0.345 $\pm$ 0.022 & 0.333 $\pm$ 0.001
& 0.332 $\pm$ 0.000 & 0.332 $\pm$ 0.000 & 0.350 $\pm$ 0.030
& 0.332 $\pm$ 0.000 & 0.332 $\pm$ 0.000 & 0.339 $\pm$ 0.012
& 0.345 $\pm$ 0.022 \\

Gemma-3-27B-it
& 0.402 $\pm$ 0.012 & 0.643 $\pm$ 0.100 & 0.681 $\pm$ 0.024
& 0.645 $\pm$ 0.027 & 0.739 $\pm$ 0.034 & 0.729 $\pm$ 0.107
& 0.472 $\pm$ 0.037 & 0.479 $\pm$ 0.087 & 0.384 $\pm$ 0.016
& 0.651 $\pm$ 0.109 \\

Tiny-Aya-Global
& 0.332 $\pm$ 0.000 & 0.340 $\pm$ 0.014 & 0.332 $\pm$ 0.000
& 0.332 $\pm$ 0.000 & 0.333 $\pm$ 0.001 & 0.376 $\pm$ 0.076
& 0.332 $\pm$ 0.000 & 0.332 $\pm$ 0.000 & 0.342 $\pm$ 0.017
& 0.347 $\pm$ 0.026 \\

\midrule
\multicolumn{11}{l}{\textbf{AfrIFact-Health -- Without Evidence (0-shot)}} \\
AfriqueQwen-14B
& 0.334 $\pm$ 0.002 & 0.333 $\pm$ 0.001 & 0.338 $\pm$ 0.005
& 0.333 $\pm$ 0.001 & 0.334 $\pm$ 0.001 & 0.333 $\pm$ 0.001
& 0.333 $\pm$ 0.001 & 0.333 $\pm$ 0.002 & 0.335 $\pm$ 0.004
& 0.333 $\pm$ 0.001 \\

Gemma-3-27B-it
& 0.717 $\pm$ 0.100 & 0.678 $\pm$ 0.117 & 0.637 $\pm$ 0.077
& 0.597 $\pm$ 0.080 & 0.702 $\pm$ 0.111 & 0.744 $\pm$ 0.113
& 0.442 $\pm$ 0.068 & 0.446 $\pm$ 0.072 & 0.452 $\pm$ 0.041
& 0.649 $\pm$ 0.098 \\

Tiny-Aya-Global
& 0.333 $\pm$ 0.001 & 0.332 $\pm$ 0.000 & 0.332 $\pm$ 0.001
& 0.333 $\pm$ 0.001 & 0.333 $\pm$ 0.001 & 0.333 $\pm$ 0.001
& 0.333 $\pm$ 0.001 & 0.333 $\pm$ 0.001 & 0.330 $\pm$ 0.003
& 0.332 $\pm$ 0.001 \\

\midrule
\multicolumn{11}{l}{\textbf{AfrIFact-Culture-News -- With Evidence (0-shot)}} \\
AfriqueQwen-14B
& 0.335 $\pm$ 0.000 & 0.376 $\pm$ 0.048 & 0.318 $\pm$ 0.004
& 0.328 $\pm$ 0.000 & 0.375 $\pm$ 0.001 & 0.366 $\pm$ 0.004
& 0.306 $\pm$ 0.000 & 0.386 $\pm$ 0.000 & 0.340 $\pm$ 0.036
& 0.368 $\pm$ 0.015 \\

Gemma-3-27B-it
& 0.806 $\pm$ 0.003 & 0.709 $\pm$ 0.139 & 0.739 $\pm$ 0.018
& 0.692 $\pm$ 0.014 & 0.741 $\pm$ 0.043 & 0.653 $\pm$ 0.117
& 0.598 $\pm$ 0.025 & 0.484 $\pm$ 0.042 & 0.642 $\pm$ 0.107
& 0.674 $\pm$ 0.096 \\

Tiny-Aya-Global
& 0.335 $\pm$ 0.000 & 0.439 $\pm$ 0.062 & 0.321 $\pm$ 0.003
& 0.328 $\pm$ 0.000 & 0.375 $\pm$ 0.000 & 0.408 $\pm$ 0.068
& 0.306 $\pm$ 0.000 & 0.386 $\pm$ 0.000 & 0.392 $\pm$ 0.124
& 0.423 $\pm$ 0.076 \\

\midrule
\multicolumn{11}{l}{\textbf{AfrIFact-Culture-News -- Without Evidence (0-shot)}} \\
AfriqueQwen-14B
& 0.353 $\pm$ 0.024 & 0.395 $\pm$ 0.014 & 0.330 $\pm$ 0.017
& 0.344 $\pm$ 0.028 & 0.383 $\pm$ 0.010 & 0.369 $\pm$ 0.002
& 0.310 $\pm$ 0.008 & 0.387 $\pm$ 0.002 & 0.322 $\pm$ 0.004
& 0.380 $\pm$ 0.007 \\

Gemma-3-27B-it
& 0.599 $\pm$ 0.018 & 0.684 $\pm$ 0.010 & 0.587 $\pm$ 0.013
& 0.601 $\pm$ 0.016 & 0.663 $\pm$ 0.041 & 0.648 $\pm$ 0.016
& 0.510 $\pm$ 0.012 & 0.438 $\pm$ 0.009 & 0.540 $\pm$ 0.014
& 0.625 $\pm$ 0.018 \\

Tiny-Aya-Global
& 0.335 $\pm$ 0.000 & 0.404 $\pm$ 0.001 & 0.317 $\pm$ 0.001
& 0.328 $\pm$ 0.000 & 0.375 $\pm$ 0.000 & 0.368 $\pm$ 0.000
& 0.306 $\pm$ 0.000 & 0.386 $\pm$ 0.000 & 0.320 $\pm$ 0.000
& 0.375 $\pm$ 0.000 \\

\bottomrule
\end{tabular}%
}
\caption{Zero-shot fact-checking accuracy prompt sensetivity scores for selected models for with and without evidence experiments across African languages in \afrifactculture and \afrifacthealth. Values are reported as mean $\pm$ standard deviation.}
\label{tab:zeroshot_evidence}
\end{table*}

}

\begin{figure*}
    \caption{\textbf{Why is evidence not helping improve accuracy?} Evidence introduces a conservative shift in model predictions: in Health, it reduces hallucinated \supports~but increases \NEI~predictions, while in Culture, it significantly improves \textsc{NEI} detection by reducing false \textsc{Supports} classifications. In the Health domain \NEI~and \supports~ improve while in the culture domain, the model tends to be good at identifying \NEI~when evidence is provided. In both domains, the model confusion between \supports~and \refutes~ improves with evidence provided.}
    \centering
    \includegraphics[width=0.8\linewidth]{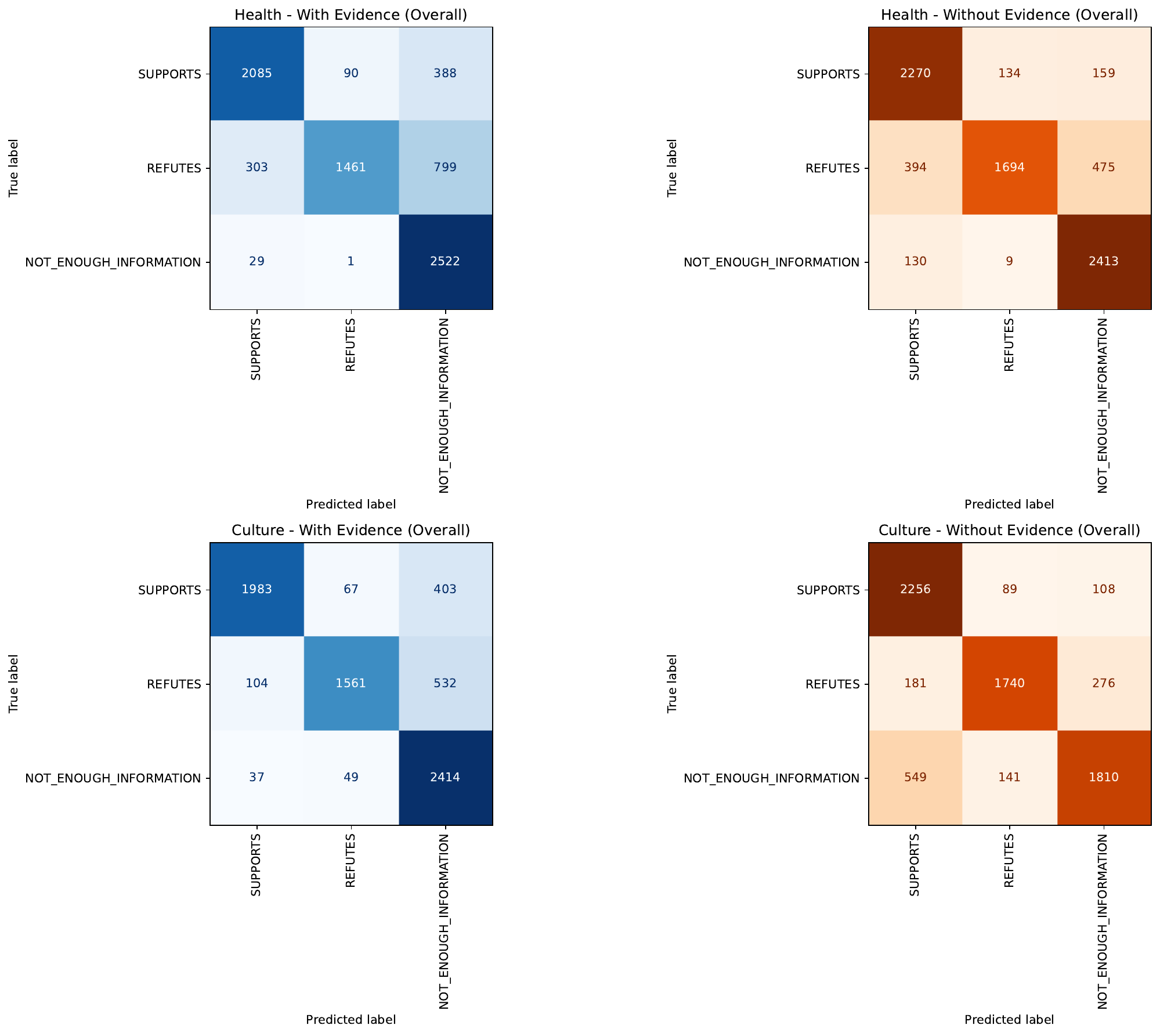}
    \label{fig:placeholder}
\end{figure*}

\section{Translation Verification Tasks}
In \afrifacthealth, all claim generation, evidence extraction, and labeling are performed in English. This design choice offers three main advantages. First, it ensures that new annotators can more easily understand and perform the task. Second, given the availability of sentence-level translations in AfriDocMT, evidence spans annotated in English can be directly translated into other languages, significantly reducing the human effort required to transfer evidence across languages and helping us cover the healthcare domain, which lacks sufficient native resources. Figure~\ref{fig:translation_comet} shows translation qualities and modifications and corrections made by translators.
\begin{figure*}[]
\caption{
Distribution of \textsc{COMET} scores for claim and document translations across African languages. 
The upper  grid shows histograms of \textsc{COMET} scores for claim translations across ten languages, 
while the lower grid presents score distributions for document translations in five languages 
(Igbo, Oromo, Shona, Twi, and Wolof).  This fixed or false alarm percentages items that has below 0.6 COMET scores colored red and addressed by annotators.
}
\centering
\includegraphics[width=\textwidth]{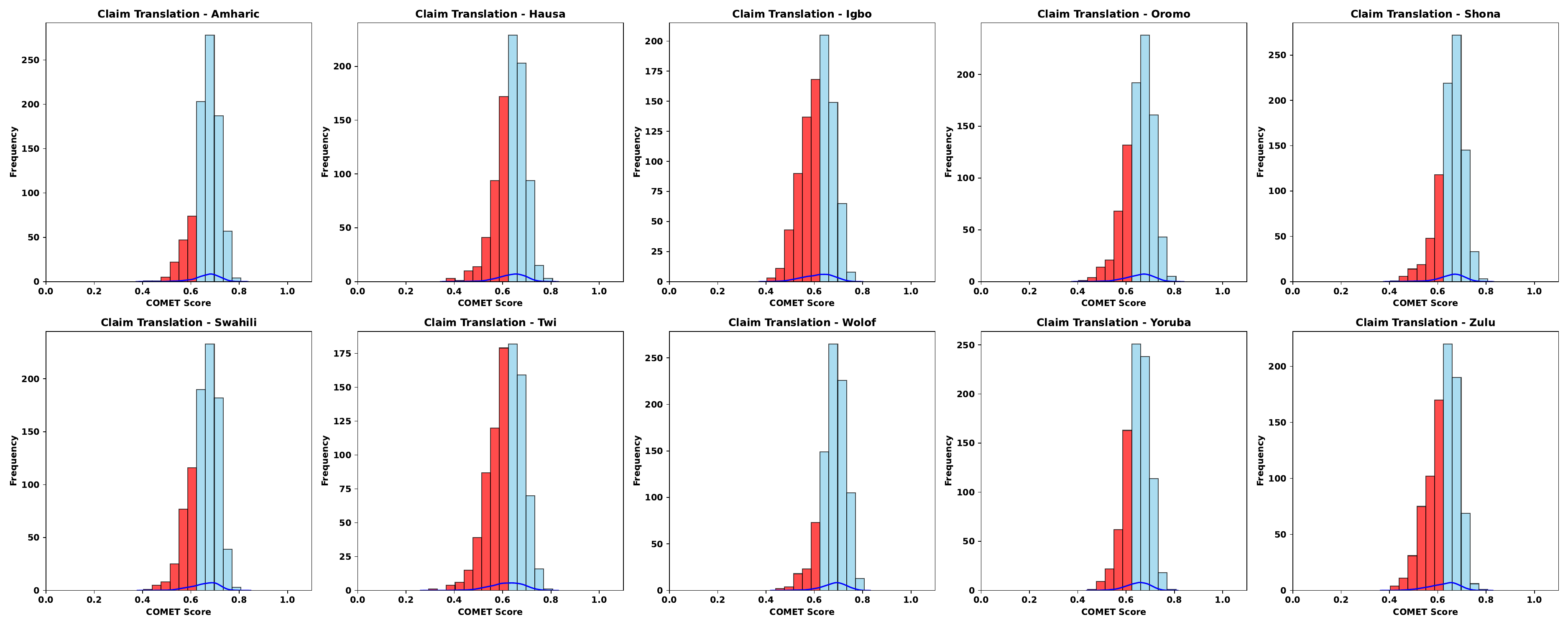}

\label{fig:translation_comet}
\end{figure*}

\begin{figure*}[!]
\caption{
Distribution of \textsc{COMET} scores for document translations across African languages.   This fixed or false alarm percentages items that has below 0.6 COMET scores colored red and addressed by annotators.
}
\centering
\includegraphics[width=\textwidth]{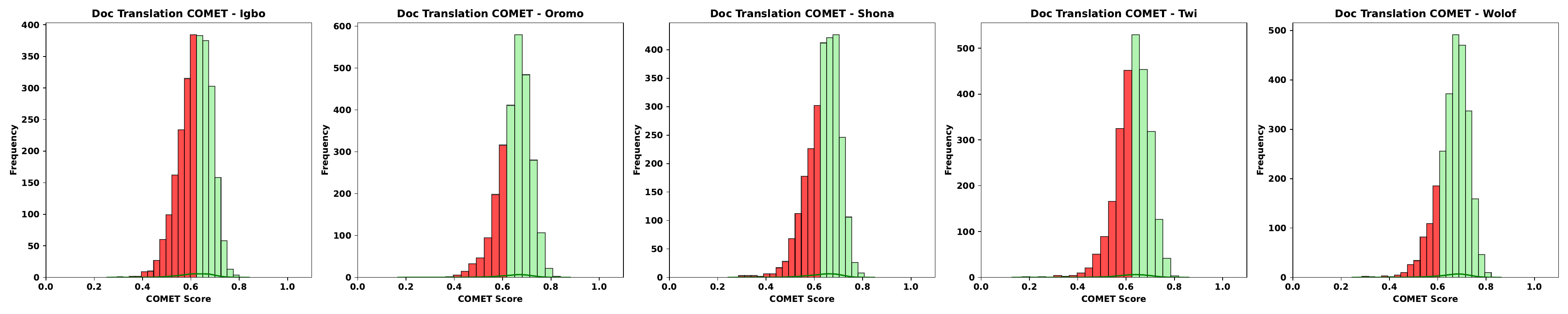}

\label{fig:translation_comet}
\end{figure*}

\section{Setup  and Reproducibility}
\label{reproduce}

\subsection{Evaluation}
We conducted all evaluations using EleutherAI’s open-source Language Model Evaluation Harness (\texttt{lm-eval})~\cite{eval-harness}. The framework supports multiple evaluation strategies, including log-likelihood, with experiments configured and managed via configuration files. For open-source models, we employed log-likelihood and generation-based evaluation for multiple-choice tasks as shown in~\cite{adelani2025irokobench}. In multiple-choice, each candidate option is appended to the corresponding question prompt, after which the log-likelihood is computed. Model accuracy is then reported based on the option with the highest log-likelihood score. For the \gptfive model, we created the same prompts and used batched API generation to reduce costs by selecting only the best prompt. Evaluation costs are close to 400 USD for GPT models across all variants. All experiments use temperature=0, top\_p=1.

\subsection{Finetuning}
To fine-tune LLLMs, we leveraged \textbf{LLaMA-Factory}~\cite{zheng2024llamafactory}, an open-source framework that supports efficient fine-tuning through parameter-efficient methods and scalable training pipelines. We performed LoRA-based supervised fine-tuning on our dataset using one H100 GPU. Our Training split was converted to the Alpaca instruction format and will be shared as part of our released resources. Training was conducted for three epochs using a LoRA rank of 8 and an effective batch size of 8 via gradient accumulation. For the QLoRA trainings we did, we kept the same hyper parameters with LORA and modified it to have 8 and 4-bit bits and bytes quantization. For full finetuning, we leveraged LLaMA-Factory, configurations with the same 3 epochs across 4 H100 GPUs. Detailed training configuration will be released in our GitHub training code examples. A systematic exploration and optimization of hyperparameters, including LoRA configuration and training dynamics, is left for future work.

\section{ Without Evidence and With Evidence Prompt Examples}

We worked on prompts that focus on $docment + claim $ for without evidence and $document + evidence + claim$ for with evidence span. Given the log-likelihood eval we used for open models, we make the claim and evidence closer together. 

{\footnotesize

\begin{tcolorbox}[
    enhanced,
    colback=gray!5!white,
    colframe=black!55,
    title=\textbf{Prompt 1},
    fonttitle=\bfseries,
    arc=2mm,
    boxrule=0.5pt,
    left=2mm,
    right=2mm,
    top=2mm,
    bottom=2mm,  
    breakable,
    fonttitle=\bfseries,
    fontupper=\footnotesize
]

\textbf{SYSTEM:} \\
You are a helpful assistant for automated fact-checking. Your task is to analyze claims based solely on the provided document.

\vspace{0.5em}

\textbf{USER:} \\
You are an intelligent decision support system designed for automated fact-checking. Determine whether the document \textbf{Supports}, \textbf{Refutes}, or provides \textbf{Not\_Enough\_Information} for the given claim.

\vspace{0.5em}

Veracity: Support, Refute, Not\_Enough\_Information,
  
Justification: Detailed reasoning addressing clarity,relevance, consistency, and sufficiency of the document

\vspace{0.5em}

\textbf{Definitions for Veracity Labels:}
\begin{itemize}
    \item \textbf{Support:} The claim is accurate and there is nothing significant missing.
    \item \textbf{Refute:} The claim is inaccurate, contradicted by the document, or makes an incorrect assertion.
    \item \textbf{Not\_Enough\_Information:} The document is insufficient, unrelated, or does not provide enough relevant information to determine whether the claim is true or false.
\end{itemize}

\vspace{0.5em}

Determine whether the claim is \textbf{Supported}, \textbf{Refuted}, or \textbf{Not\_Enough\_Information} based \textbf{ONLY} on the provided document.

\vspace{0.5em}

\textbf{Document:} \\
\{DOCUMENT\}

\vspace{0.5em}

\textbf{Claim:} \\
\{CLAIM\}

\vspace{0.5em}

\textbf{Question:} \\
Based only on the document above, does it Support, Refute, or provide Not\_Enough\_Information for the claim?

\vspace{0.5em}

\textbf{Answer:} (choose \supports, \refutes, or \NEI)

\end{tcolorbox}
}
{\footnotesize
\begin{tcolorbox}[
    enhanced,
    colback=gray!5!white,
    colframe=black!55,
    title=\textbf{Prompt 2},
    fonttitle=\bfseries,
    arc=2mm,
    boxrule=0.5pt,
    left=2mm,
    right=2mm,
    top=2mm,
    bottom=2mm,
    breakable
]

\textbf{Role:} \\
You are a professional fact checker responsible for verifying factual claims.

\vspace{0.5em}

\textbf{Objective:} \\
Assess the relationship between a claim and the provided document by determining whether the document \textbf{Supports}, \textbf{Refutes}, or provides \textbf{Not\_Enough\_Information} for the claim.

\vspace{0.5em}

\textbf{Constraints:}
\begin{itemize}
    \item Use only the document provided below.
    \item Do not rely on prior knowledge, assumptions, or external information.
    \item If the document does not clearly support or contradict the claim, select \textbf{Not\_Enough\_Information}.
\end{itemize}

\vspace{0.5em}

\textbf{Document:} \\
\{DOCUMENT\}

\vspace{0.5em}

\textbf{Claim:} \\
\{CLAIM\}

\vspace{0.5em}

\textbf{Question:} \\
Based only on the document above, does it \textbf{Support}, \textbf{Refute}, or provide \textbf{Not\_Enough\_Information} for the claim?

\vspace{0.5em}

\textbf{Answer:} (choose \supports, \refutes, or \NEI)

\end{tcolorbox}

}
{\scriptsize

\begin{tcolorbox}[
    enhanced,
    colback=gray!5!white,
    colframe=black!55,
    title=\textbf{Prompt 3},
    fonttitle=\bfseries,
    arc=2mm,
    boxrule=0.5pt,
    left=2mm,
    right=2mm,
    top=2mm,
    bottom=2mm,
    breakable
]

\textbf{SYSTEM:} \\
You are a helpful assistant for Multilingual Evidence-Centered Fact Verification. Your task is to analyze claims based solely on the provided documents.

\vspace{0.5em}

\textbf{USER:} \\
You are an intelligent decision support system designed for automated fact-checking. Determine whether the document \textbf{Supports}, \textbf{Refutes}, or provides \textbf{Not\_Enough\_Information} for the given claim.

\vspace{0.5em}

\textbf{Role:} \\
You are an independent fact checker tasked with evaluating factual claims in linguistically diverse settings.

\vspace{0.5em}

\textbf{Objective:} \\
Determine the factual status of a claim by interpreting the provided document, which may appear in different languages or linguistic varieties, and decide whether the document confirms the claim, contradicts it, or fails to address it.

\vspace{0.5em}

\textbf{Constraints:}
\begin{itemize}
    \item Base your judgment exclusively on the document provided, regardless of language or linguistic variation.
    \item Do not use background knowledge, assumptions, or external sources.
    \item If the document does not directly confirm or contradict the claim, choose \textbf{Not\_Enough\_Information}.
\end{itemize}

\vspace{0.5em}

\textbf{Document:} \\
\{DOCUMENT\}

\vspace{0.5em}

\textbf{Claim:} \\
\{CLAIM\}

\vspace{0.5em}

\textbf{Question:} \\
Based only on the document above, does it \textbf{Support}, \textbf{Refute}, or provide \textbf{Not\_Enough\_Information} for the claim?

\vspace{0.5em}

\textbf{Answer:} (choose \supports, \refutes, or \NEI)

\end{tcolorbox}
}
{\scriptsize
\begin{tcolorbox}[
    enhanced,
    colback=gray!5!white,
    colframe=black!55,
    title=\textbf{Prompt 1},
    fonttitle=\bfseries,
    arc=2mm,
    boxrule=0.5pt,
    left=2mm,
    right=2mm,
    top=2mm,
    bottom=2mm,
    breakable
]

\textbf{SYSTEM:} \\
You are a helpful assistant for automated fact-checking. Your task is to analyze claims based on the provided evidence.

\vspace{0.5em}

\textbf{USER:} \\
You are an intelligent decision support system designed for automated fact-checking. Based \textbf{only} on the evidence provided below, determine whether the evidence \textbf{Supports}, \textbf{Refutes}, or provides \textbf{Not\_Enough\_Information} for the given claim.

\vspace{0.5em}

Veracity: Support, Refute, Not\_Enough\_Information

Justification: Detailed reasoning addressing clarity, relevance, consistency, and sufficiency of the evidence

\vspace{0.5em}

\textbf{Definitions for Veracity Labels:}
\begin{itemize}
    \item \textbf{Support:} The claim is accurate and there is nothing significant missing.
    \item \textbf{Refute:} The claim is inaccurate, contradicted by the evidence, or makes an incorrect assertion.
    \item \textbf{Not\_Enough\_Information:} The evidence is insufficient, unrelated, or does not provide enough relevant information to determine whether the claim is true or false.
\end{itemize}

\vspace{0.5em}

Determine whether the claim is \textbf{Supported}, \textbf{Refuted}, or \textbf{Not\_Enough\_Information} based \textbf{ONLY} on the provided evidence.

\vspace{0.5em}

\textbf{Document:} \\
\{DOCUMENT\}

\vspace{0.5em}

\textbf{Evidence:} \\
\{EVIDENCE\}

\vspace{0.5em}

\textbf{Claim:} \\
\{CLAIM\}

\vspace{0.5em}

\textbf{Question:} \\
Based only on the evidence above, does the evidence Support, Refute, or provide Not\_Enough\_Information for the claim?

\vspace{0.5em}

\textbf{Answer:} (choose \supports, \refutes, or \NEI)

\end{tcolorbox}
}
{\scriptsize
\begin{tcolorbox}[
    enhanced, 
    colback=gray!5!white,
    colframe=black!55,
    title=\textbf{Prompt 2},
    fonttitle=\bfseries,
    arc=2mm,
    boxrule=0.5pt,
    left=2mm,
    right=2mm,
    top=2mm,
    bottom=2mm,
    breakable
]

\textbf{Role:} \\
You are a professional fact checker responsible for verifying factual claims.

\vspace{0.5em}

\textbf{Objective:} \\
Assess the relationship between a claim and the provided evidence by determining whether the evidence \textbf{Supports}, \textbf{Refutes}, or provides \textbf{Not\_Enough\_Information} for the claim.

\vspace{0.5em}

\textbf{Constraints:}
\begin{itemize}
    \item Use only the evidence provided below.
    \item Do not rely on prior knowledge, assumptions, or external information.
    \item If the evidence does not clearly support or contradict the claim, select \textbf{Not\_Enough\_Information}.
\end{itemize}

\vspace{0.5em}

\textbf{Document:} \\
\{DOCUMENT\}

\vspace{0.5em}

\textbf{Evidence:} \\
\{EVIDENCE\}

\vspace{0.5em}

\textbf{Claim:} \\
\{CLAIM\}

\vspace{0.5em}

\textbf{Question:} \\
Based only on the evidence above, does the evidence \textbf{Support}, \textbf{Refute}, or provide \textbf{Not\_Enough\_Information} for the claim?

\vspace{0.5em}

\textbf{Answer:} (choose \supports, \refutes, or \NEI)

\end{tcolorbox}

}
{\scriptsize
\begin{tcolorbox}[
    enhanced,
    colback=gray!5!white,
    colframe=black!55,
    title=\textbf{Prompt 3},
    fonttitle=\bfseries,
    arc=2mm,
    boxrule=0.5pt,
    left=2mm,
    right=2mm,
    top=2mm,
    bottom=2mm,
    breakable
]

\textbf{SYSTEM:} \\
You are a helpful assistant for Multilingual Evidence-Centered Fact Verification. Your task is to analyze claims based on the provided evidence.

\vspace{0.5em}

\textbf{USER:} \\
You are an intelligent decision support system designed for automated fact-checking. Based only on the evidence provided below, determine whether the evidence \textbf{Supports}, \textbf{Refutes}, or provides \textbf{Not\_Enough\_Information} for the given claim.

\vspace{0.5em}

\textbf{Role:} \\
You are an independent fact checker tasked with evaluating factual claims in linguistically diverse settings.

\vspace{0.5em}

\textbf{Objective:} \\
Determine the factual status of a claim by interpreting the provided evidence, which may appear in different languages or linguistic varieties, and decide whether the evidence confirms the claim, contradicts it, or fails to address it.

\vspace{0.5em}

\textbf{Constraints:}
\begin{itemize}
    \item Base your judgment exclusively on the evidence provided, regardless of language or linguistic variation.
    \item Do not use background knowledge, assumptions, or external sources.
    \item If the evidence does not directly confirm or contradict the claim, choose \textbf{Not\_Enough\_Information}.
\end{itemize}

\vspace{0.5em}

\textbf{Document:} \\
\{DOCUMENT\}

\vspace{0.5em}

\textbf{Evidence:} \\
\{EVIDENCE\}

\vspace{0.5em}

\textbf{Claim:} \\
\{CLAIM\}

\vspace{0.5em}

\textbf{Question:} \\
Based only on the evidence above, does the evidence \textbf{Support}, \textbf{Refute}, or provide \textbf{Not\_Enough\_Information} for the claim?

\vspace{0.5em}

\textbf{Answer:} (choose \supports, \refutes, or \NEI)

\end{tcolorbox}

}

\FloatBarrier
\section{Annotation Tool Examples}
\label{annotation-tool}

\begin{figure}[!]
    \centering
    \includegraphics[width=\linewidth]{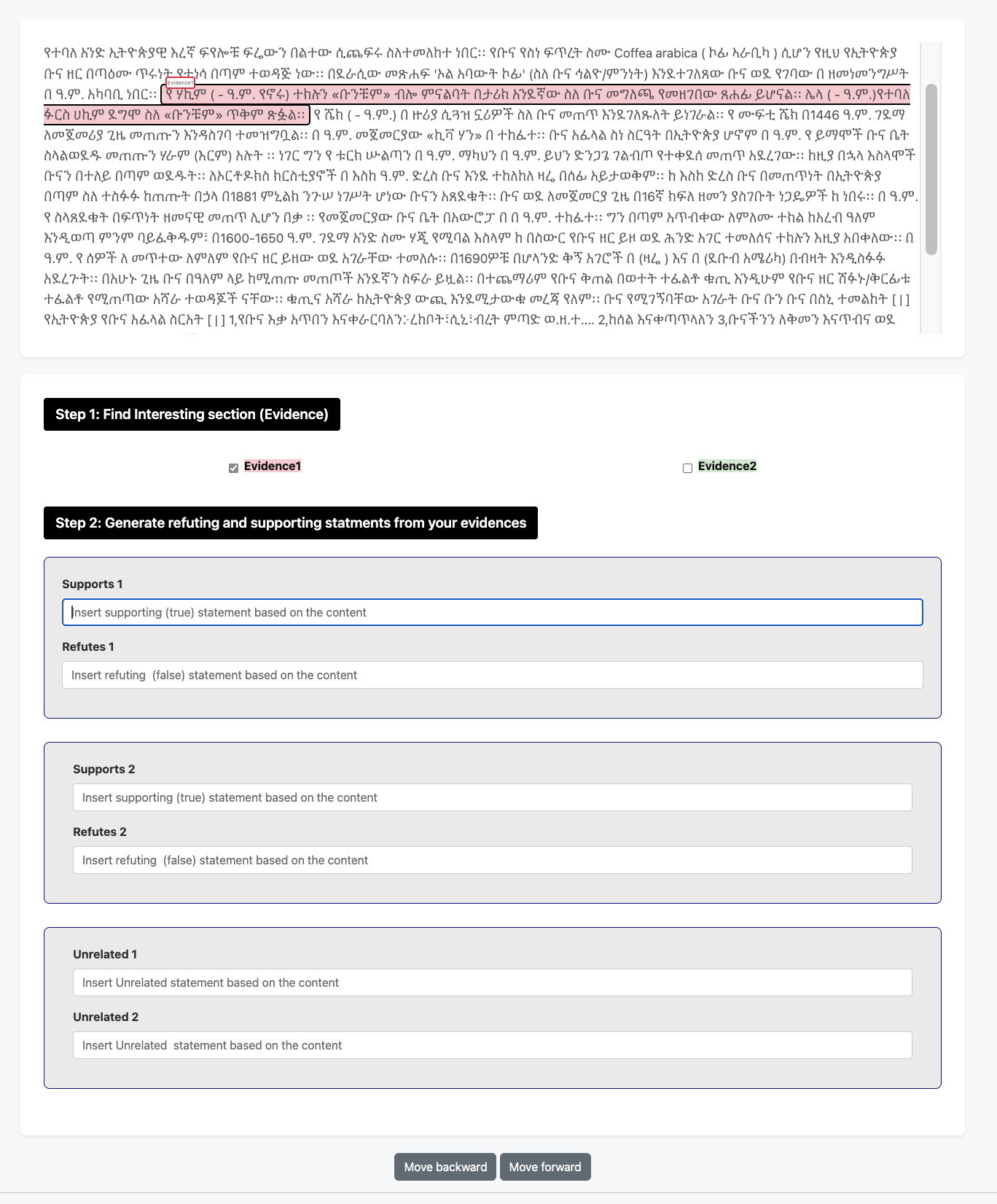}
    \caption{Example interface of the customized tool used for claim generation, showing an Amharic example.}
    \label{fig:placeholder1}
\end{figure}

\begin{figure}[!]
    \centering
    \includegraphics[width=\linewidth]{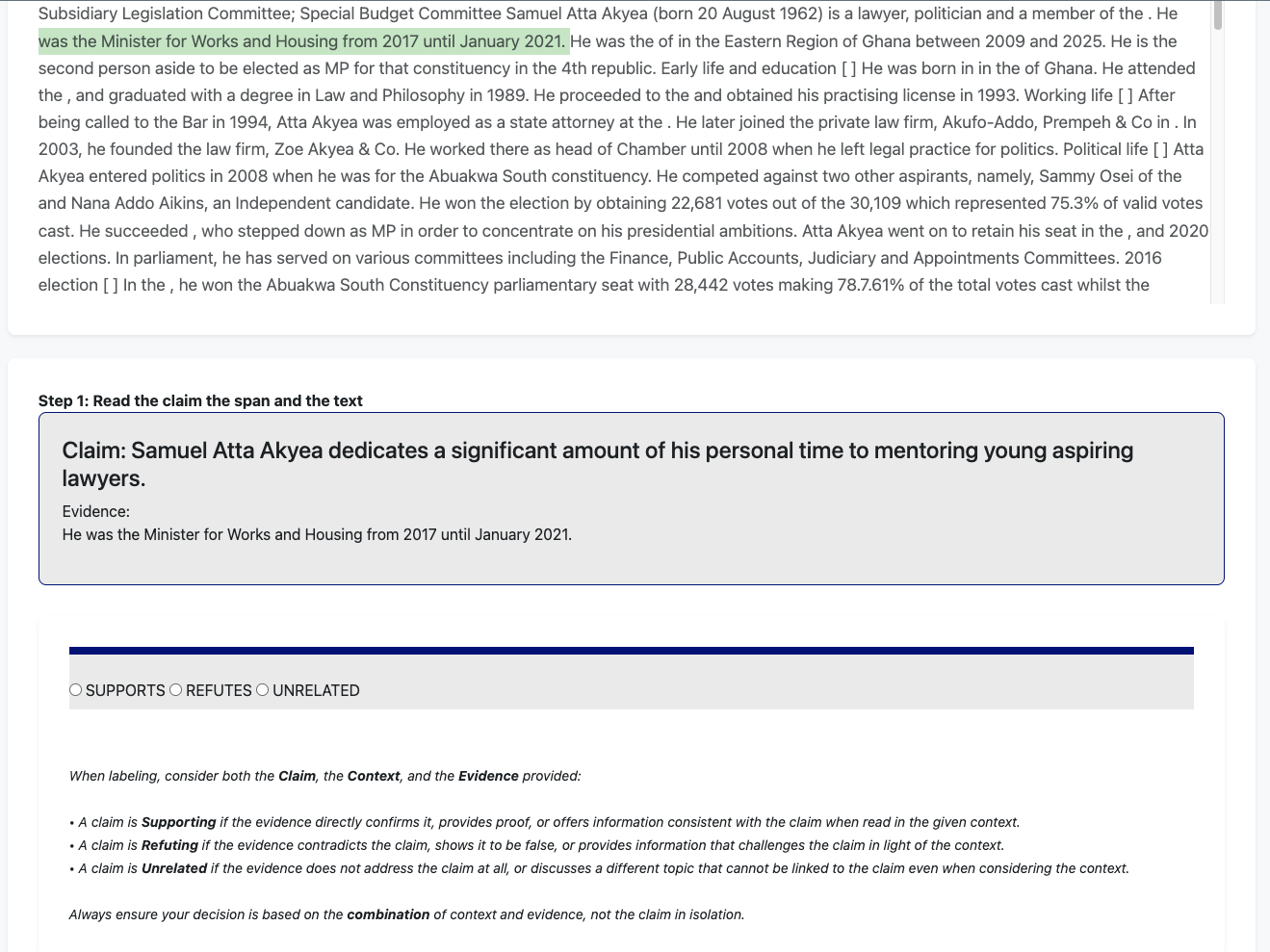}
    \caption{Example of the customized annotation interface used for labeling claims in the claim labeling, showing an English example.}
    \label{fig:placeholder2}
\end{figure}

Figures~\ref{fig:placeholder1} and ~\ref{fig:placeholder2} show screenshots of tools used for claim generation and claim labeling tasks.

\FloatBarrier
\section{Annotation Guide}
Due to multiple tasks and multiple platforms used in our task, we leveraged several guides, including video, textual, and similar practices on how to perform the tasks, followed by tests and back-and-forth with language coordinators to verify the validity of the tasks. 


\begin{tcolorbox}[
title=\textbf{AfriFact Annotation Guidelines},
colback=gray!5,
colframe=black!70,
boxrule=0.5pt,
breakable,
fonttitle=\bfseries,
fontupper=\footnotesize
]

\textbf{Task 1: Claim Generation}

\textbf{Claim Definition:}
A claim is a single sentence expressing information about one target entity.

\textbf{Requirements}
\begin{itemize}[leftmargin=*]
\item Must reference the main entity directly (no pronouns).
\item Avoid vague expressions (e.g., “may be”, “it is reported”).
\item Claims must be derivable from the source sentence and dictionary.
\item Include temporal context when facts may change.
\end{itemize}

\textbf{Evidence Labels}
\begin{itemize}[leftmargin=*]
\item \textbf{SUPPORTED}: Evidence agrees with the claim.
\item \textbf{REFUTED}: Evidence contradicts the claim.
\item \textbf{NOT ENOUGH INFO}: Evidence insufficient to determine truth.
\end{itemize}

\textbf{Task 2: Claim Labeling}

\textbf{Objective:}
Label claims and select the minimal evidence needed.

\textbf{Procedure}
\begin{enumerate}[leftmargin=*]
\item Read the claim and candidate evidence sentences.
\item Determine if evidence supports or refutes the claim.
\item Combine multiple sentences if needed.
\item If no sufficient evidence exists, label \textit{NOT ENOUGH INFO}.
\end{enumerate}

\textbf{Rule of Thumb}

\textit{If only the selected evidence is given, can the claim be verified as true or false? If not, label NOT ENOUGH INFO.}

\textbf{Edge Case Rules}
\begin{itemize}[leftmargin=*]
\item Avoid ambiguous claims (e.g., many, several, popular).
\item Distinguish between actors and fictional characters.
\item Filmographies and lists are not exhaustive evidence.
\item Be time-aware when verifying roles or titles.
\end{itemize}

\textbf{Skipping Rules}
\begin{itemize}[leftmargin=*]
\item Claim cannot be verified from Wikipedia.
\item Claim is vague or personal.
\item Claim contains major grammatical errors.
\end{itemize}

\end{tcolorbox}


\end{document}